\documentclass[lettersize,journal]{IEEEtran}
\usepackage{amsmath,amsfonts}
\usepackage{algorithmic}
\usepackage{array}
\usepackage[caption=false,font=normalsize,labelfont=sf,textfont=sf]{subfig}
\usepackage{textcomp}
\usepackage{stfloats}
\usepackage{url}
\usepackage{verbatim}
\usepackage{bm}
\usepackage{graphicx}
\usepackage{multirow}
\usepackage{amssymb}
\usepackage{float}
\usepackage{subfig}
 \usepackage{booktabs} 
 \usepackage{marvosym}

\title{
StreetSurfGS: Scalable Urban Street Surface Reconstruction with Planar-based Gaussian Splatting
}

\author{
Xiao Cui$^*$\thanks{Xiao Cui, Wengang Zhou and
Houqiang Li are with the Department of Electrical Engineering
and Information Science, University of Science and Technology of
China, Hefei, 230027 China (e-mail: cuixiao2001@mail.ustc.edu.cn,
zhwg@ustc.edu.cn, lihq@ustc.edu.cn)}\thanks{Weicai Ye and Guofeng Zhang are with the State Key Laboratory of CAD and CG, Zhejiang University, Hangzhou 310058, China
(e-mail: weicaiye@zju.edu.cn, zhangguofeng@zju.edu.cn).
}
\thanks{Yifan Wang is with Shanghai AI Laboratory, Shanghai 200233, China (e-mail: wangyifan@pjlab.org.cn)}, Weicai Ye$^*$$^\textrm{\Letter}$\thanks{$^*$: Equal Contribution. \textrm{\Letter}: Corresponding Author.}, Yifan Wang, Guofeng Zhang~\IEEEmembership{Member,~IEEE,
} Wengang Zhou,~\IEEEmembership{Senior Member,~IEEE,
} %Tong He$^\textrm{\Letter}$, 
Houqiang Li,~\IEEEmembership{Fellow,~IEEE} }
\IEEEoverridecommandlockouts

\begin{document}
% \markboth{IEEE TRANSACTIONS ON CIRCUITS AND SYSTEMS FOR VIDEO TECHNOLOGY
% , 2024}%
% {Shell \MakeLowercase{\etal}: A Sample Article Using IEEEtran.cls for IEEE Journals}
\maketitle

\begin{figure*}[h]
%缺小地图
\centering
 \includegraphics[width=1\textwidth]{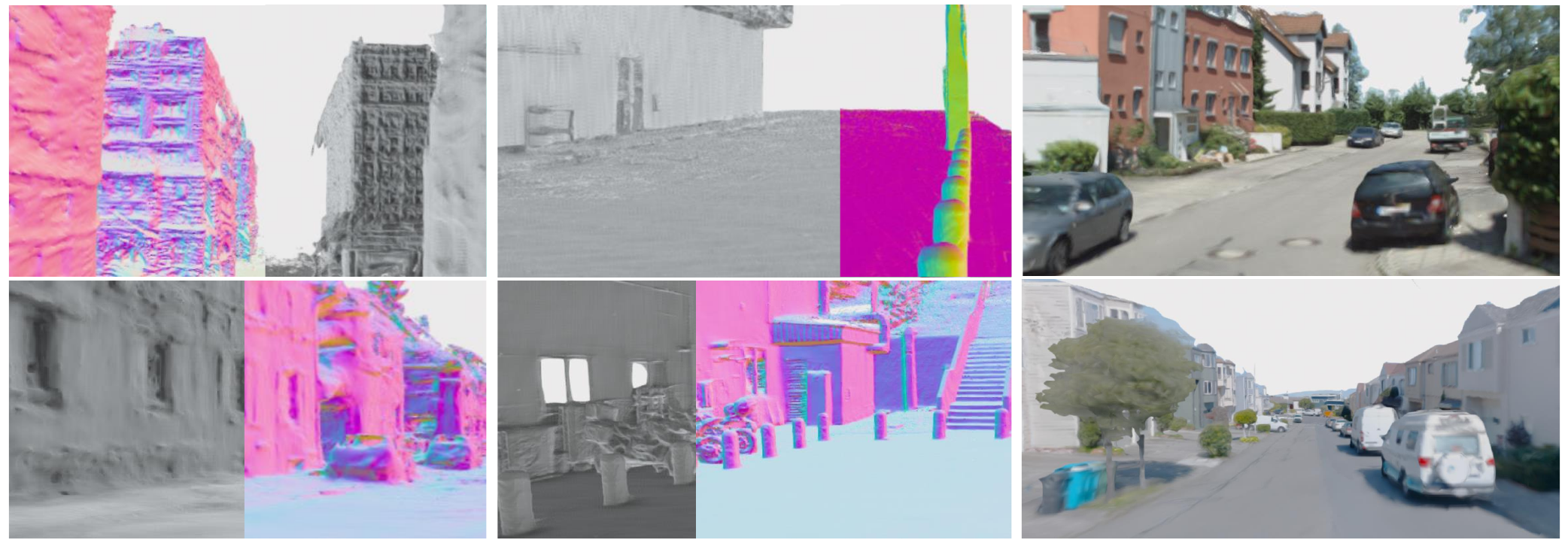}
\caption{We present StreetSurfGS, a framework for scalable street scenes surface reconstruction from RGB images using Gaussian Splatting, without the need of lidar or pretrained geometry estimation models for supervision. Shown in the figure are untextured meshes, normals and textured meshes.}\label{teaser}
\vspace{-0pt}
\end{figure*}

\begin{abstract}
%重写
% 先介绍街景重建很重要，具有那三个特点；然后说3dgs的surface reconstruction方法都是为object-centric场景设计的，不能很好配适街景；再说我们是第一个用3dgs做街景的；我们提出一系列方法分别配适这三个特点；最后说实验表明novel view synthesis和surface reconstruction都能做好
\vspace{-2pt}
Reconstructing urban street scenes is crucial due to its vital role in applications such as autonomous driving and urban planning. These scenes are characterized by long and narrow camera trajectories, occlusion, complex object relationships, and data sparsity across multiple scales. Despite recent advancements, existing surface reconstruction methods, which are primarily designed for object-centric scenarios, struggle to adapt effectively to the unique characteristics of street scenes.
To address this challenge, we introduce StreetSurfGS, the first method to employ Gaussian Splatting specifically tailored for scalable urban street scene surface reconstruction. StreetSurfGS utilizes a planar-based octree representation and segmented training to reduce memory costs, accommodate unique camera characteristics, and ensure scalability. Additionally, to mitigate depth inaccuracies caused by object overlap, we propose a guided smoothing strategy within regularization to eliminate inaccurate boundary points and outliers.
Furthermore, to address sparse views and multi-scale challenges, we use a dual-step matching strategy that leverages adjacent and long-term information. Extensive experiments validate the efficacy of StreetSurfGS in both novel view synthesis and surface reconstruction.


\end{abstract}

\section{Introduction}
\IEEEPARstart{T}{he} reconstruction of large-scale street scenes is critically important for applications ranging from autonomous driving~\cite{ye2023pvo, ye2022deflowslam} to urban planning. While significant advancements have been made in novel view synthesis, the challenge of surface reconstruction has received comparatively less attention. Previous methods such as DNMP~\cite{lu2023urban} and StreetSurf~\cite{guo2023streetsurf} employ NeRF-based techniques for this purpose. However, these methods either rely on a complete mesh representation or require an entire volume for marching cubes, limiting their scalability for large scenes. Recent advancements in 3D Gaussian Splatting (3DGS)~\cite{kerbl20233d} have revolutionized novel view synthesis, boasting high fidelity and exceptional training and rendering speeds~\cite{yu2023mip,yan2023multi,liang2024analytic,lu2023scaffold,cheng2024gaussianpro,ren2024octree}. 
However, when applied to the reconstruction of large outdoor scenes, 3DGS exhibits notable shortcomings, often resulting in poor geometric accuracy and high memory consumption~\cite{guedon2023sugar,jiang2023gaussianshader}.
These limitations highlight a significant gap and underscore the potential for solutions that balance robustness with computational efficiency.

While some recent advancements have improved surface reconstruction capabilities~\cite{guedon2023sugar,chen2024pgsr},
their reliance on the traditional 3DGS representation often leads to excessive memory usage in large scenes, thereby limiting their practical utility in real-world applications.
Furthermore, existing methods are primarily designed for object-centric scenarios and are not suitable for street view scenes. 
Street views present inherent challenges due to their long and narrow camera trajectories, necessitating the processing of extensive street view data and resulting in high computational complexity. Additionally, the occlusion and complex relationships between objects in street views significantly increase modeling difficulty.
Moreover, the continuous forward movement of the camera results in infrequent occurrences of scenes varying in size and distance from the camera. This sparsity contrasts with object-centric scenarios, leading to fewer consensus regions between frames. This further compounds the complexity of street scene reconstruction. These properties underscore the necessity for a specialized approach to achieve effective surface reconstruction in street scenes.

In response to these challenges, we introduce StreetSurfGS, the first attempt to use Gaussian Splatting for scalable street scene surface reconstruction. To address the high memory demands typically associated with large-scale urban modeling, StreetSurfGS incorporates a planar-based octree gaussian splatting representation. 
This approach significantly diminishes memory consumption by dynamically adjusting octree layers throughout the urban environment. Intricate areas such as intersections and detailed architectural features are delineated with finer octree layers, while less complex regions like open roads are characterized by coarser octree layers. By focusing resources on areas requiring higher detail, StreetSurfGS ensures that the structural fidelity of urban environments is meticulously captured and preserved.
To accommodate the long and narrow characteristics of street scenes and effectively handle the dynamic nature of the data stream, we employ a segmented training strategy that enables scalability. This approach segments the captured video frames temporally with intentional overlap between segments, ensuring continuous and coherent data processing. 

To address depth continuity errors arising from occlusion and complex relationships between objects, our approach integrates the Segment Anything Model (SAM)~\cite{kirillov2023segment}. SAM detects edges, followed by edge and outlier filtering within the smooth regularization framework, thereby enhancing depth consistency while ensuring seamless transitions within objects. This leads to accurate depth representation, accommodating both continuous and discontinuous depth changes within and between objects.
Moreover, to tackle the sparse viewpoint challenges inherent in street view scenes and manage details across varying distances, we adopt a dual strategy for multi-view consistency regularization. Firstly, we enhance local geometric and photometric consistency by imposing additional constraints from adjacent viewpoints. Secondly, we harness long-term information by establishing far-field matching constraints, leveraging consensus regions across distant frames to mitigate accumulated errors. Through cross-scale information propagation, we effectively address the issue of data sparsity across multiple scales, thereby enhancing the consistency and stability of the reconstruction.
Our approach undergoes quantitative and qualitative evaluations across four datasets. Extensive experiments demonstrate its superior rendering quality and satisfactory surface reconstruction results.
We showcase surface reconstruction results in Figure~\ref{teaser}, and summarize our contributions as follows:
% 第一个做，解决问题1， 解决问题2和3
\begin{itemize}
    \item We analyze the unique characteristics of street views and introduce StreetSurfGS, marking the first instance of applying Gaussian Splatting specifically tailored for scalable urban street scene surface reconstruction. Through comprehensive evaluation on urban datasets, our method demonstrates superiority in both novel view synthesis and geometry reconstruction.
    \item We integrate a planar-based octree Gaussian Splatting representation and employ a segmented training strategy specially tailored to the unique characteristics of street views. These enhancements facilitate the efficient reconstruction of urban environments characterized by long and narrow camera trajectories, reducing memory consumption and ensuring scalability.
     \item We employ a guided smoothing strategy within regularization to mitigate depth inaccuracy caused by object overlap. Also, we use a dual-step matching strategy that leverages both adjacent and long-term information to address sparse views and multi-scale challenges.
     % \item We integrate a planar-based octree gaussian splatting representation and a segmented training strategy to address the unique characteristics of street views. These enhancements allow for efficient reconstruction of long and narrow urban environments and ensure scalability.
   % \item To address object occlusion, we 
    
   %  To tackle sparse view characteristics inherent in street scenes, we implement a multi-view consistency mechanism. This mechanism incorporates both multiple neighboring frames and keyframe strategies, effectively balancing depth discontinuities and maintaining consistency within individual objects. This capability enables accurate reconstruction of irregular shapes and layouts encountered in urban street views.
\end{itemize}

\section{Related Work}
In this section, we briefly revisit the related topic, including large scene reconstruction, urban street scene reconstruction and Gaussian Spatting.

\subsection{Large Scene Reconstruction}
Significant advancements have been made in image-based large scene reconstruction over recent decades. To get camera positions and create a sparse 3D point cloud, multiple studies \cite{agarwal2011building, fruh2004automated, li2008modeling, pollefeys2008detailed, zhu2018very, schonberger2016structure, liu2021coxgraph} leverage a structure-from-motion (SfM) pipeline. Many works \cite{furukawa2010towards,goesele2007multi,lin2022multiview,li2023hybrid,kim2012outdoor,lee2021high,li2011markerless,su2022uncertainty} also focus on producing dense point clouds or triangle meshes %from the SfM output 
using multi-view stereo (MVS) techniques~\cite{metashape}. These methods try to estimate the depth map
by matching feature points across different views. 
For instance, Li et al. \cite{li2011markerless} introduce a marker-less shape and motion capture technique for multi-view video sequences, while Su et al. \cite{su2022uncertainty} developed an uncertainty-guided network along with an uncertainty-aware loss function to address unsupervised perception of uncertainty.
% Specifically, Li et al. \cite{li2011markerless} propose a
% marker-less shape and motion capture approach for multi-view video sequences, Su et al. \cite{su2022uncertainty} propose an uncertainty-guided network and an uncertainty-aware loss to perceive the uncertainty in an unsupervised manner.

In recent years, Neural Radiance Fields (NeRF) \cite{mildenhall2021nerf, Ye2023IntrinsicNeRF, huang2024nerf, ming2022idf} emerges as a popular 3D representation for achieving photo-realistic novel-view synthesis, as evidenced by a growing body of work \cite{barron2021mip,barron2022mip,barron2023zip,yariv2021volume,fridovich2022plenoxels,muller2022instant,chen2022tensorf,reiser2021kilonerf,yu2021plenoctrees,sun2022direct,hedman2021baking,cao2023hexplane,fridovich2023k}. Extending from NeRF, many neural surface reconstruction methods~\cite{wang2021neus, tang2024nd, wang2024neurodin, li2023neuralangelo} have emerged. Several methods \cite{tancik2022block,turki2022mega,xiangli2022bungeenerf,xu2023grid} extend NeRF to handle large-scale scenes novel-view synthesis. 
% For instance, Block-NeRF \cite{tancik2022block} partitions urban environments into multiple blocks and allocates training views based on their spatial distribution. Mega-NeRF \cite{turki2022mega} employs a grid-based approach, assigning each pixel in an image to different grids traversed by its corresponding ray. In contrast, Grid-NeRF \cite{xu2023grid} integrates NeRF and grid-based methods without scene decomposition. Additionally, VastGaussian \cite{lin2024vastgaussian} uses Gaussian splatting to partition large scenes into cells and merge the seperate scenes into a complete scene after parallel optimization. Despite these advancements enhancing rendering quality, large scene surface reconstruction remains relatively underexplored.
For instance, Block-NeRF \cite{tancik2022block} segments urban environments into manageable blocks, allocating training views based on their spatial distribution, while Mega-NeRF \cite{turki2022mega} employs a grid-based strategy, associating each pixel in an image with different grids traversed by its corresponding ray. In contrast, Grid-NeRF~\cite{xu2023grid} integrates NeRF with grid-based methods without the need for explicit scene decomposition. 

Some efforts have also been made to explore the application of 3D Gaussian Splatting in large-scale scene image rendering. VastGaussian~\cite{lin2024vastgaussian} introduces Gaussian splatting to partition large scenes into cells, subsequently merging the separate components into a unified scene after parallel optimization. Similarly, DoGaussian~\cite{chen2024dogaussian} enhances the training efficiency by decomposing scenes into blocks and employing a consensus algorithm to ensure global model consistency.
Despite these significant advancements, large-scale scene surface reconstruction remains relatively underexplored, highlighting the need for further research.

\subsection{Urban Street Scene Surface Reconstruction}
Reconstructing urban street scene surfaces presents a complex challenge due to the intricacies of camera trajectories, object interactions, and the sparse viewpoints typically available. Recent efforts have sought to address these challenges with varying approaches. DNMP~\cite{lu2023urban} introduces a flexible neural extension of the traditional mesh representation, demonstrating robust capabilities for photorealistic image synthesis. Urban NeRF~\cite{rematas2022urban} leverages LiDAR point clouds to enhance geometric learning. Unlike them, StreetSurf~\cite{guo2023streetsurf} utilizes geometric priors (e.g., depth, normal) derived from pretrained monocular models for supervision, achieving notable reconstruction quality. However, these methods are constrained by their reliance on either complete mesh representations or the need to process entire volumes for marching cubes, which hampers their scalability for large-scale scenes.

More recently, advanced methods such as Driving Gaussian~\cite{zhou2023drivinggaussian} and Street Gaussian~\cite{yan2024street} have effectively combined LiDAR data with images to enable novel view synthesis in dynamic urban environments using 3D Gaussian Splatting (3DGS). While effective, these sophisticated approaches depend heavily on additional LiDAR information for depth supervision and still lack inherent mechanisms to guarantee geometric accuracy, rendering them less suitable for high-quality surface reconstruction based solely on RGB images. In contrast, StreetSurfGS is the pioneering method for urban street scene surface reconstruction using Gaussian Splatting.%, offering scalability advantages. 
It achieve a balance between scalability and precision without the need for supplementary depth or normal information.

\subsection{Gaussian Splatting}
3D Gaussian Splatting (3DGS)~\cite{kerbl20233d} is a cutting-edge rendering technique that uses 3D Gaussians to represent and synthesize highly realistic 3D scenes directly from 2D images, bypassing traditional mesh creation methods and enabling efficient, real-time novel view synthesis.
Recently, 3DGS has seen groundbreaking advancements, with applications in areas such as quality enhancement~\cite{yu2023mip,yan2023multi,liang2024analytic,lu2023scaffold,cheng2024gaussianpro,ren2024octree, cai2024dynasurfgs}, compression and regularization ~\cite{navaneet2023compact3d,lee2023compact,girish2023eagles,niedermayr2023compressed}, and physical simulation ~\cite{wu20234d,liang2023gaufre,duan20244d,sun20243dgstream}.

In the realm of geometry reconstruction, NeuSG~\cite{chen2023neusg} integrates the previous NeRF-based surface reconstruction method NeuS~\cite{wang2021neus} into the 3DGS framework to transfer surface properties to the 3DGS. SuGaR~\cite{guedon2023sugar} extracts explicit meshes from the 3DGS representation by regularizing Gaussians over surfaces. Building on SuGaR, GaMeS~\cite{waczynska2024games}  uses these explicit meshes as input, parameterizing Gaussian components through vertices, enabling real-time modification of Gaussians by altering
mesh components during inference. 2DGS~\cite{huang20242d} simplifies the 3D volume into 2D oriented planar Gaussian disks, while GOF~\cite{yu2024gaussian} introduces a Gaussian opacity field, identifying a level set to enable precise geometry extraction. To maintain a trade-off between rendering quality and geometry reconstruction, GSDF~\cite{yu2024gsdf} combines 3DGS and SDF into a dual-branch architecture. PGSR~\cite{chen2024pgsr} employs a planer-based Gaussian representation and proposes an unbiased depth rendering method. However, these methods primarily cater to object-centric camera trajectories and are not well-suited for urban street scenes, which are typically long and narrow. Furthermore, their approaches often encounter memory limitations when handling the extensive data characteristic of urban environments. In our work, we make the first attempt to conduct scalable urban street scene surface reconstruction using Gaussian Splatting.

% Recently, 3D Gaussian Splatting has made groundbreaking advancements, with
% applications in 
% quality enhancement \cite{yu2023mip,yan2023multi,liang2024analytic,lu2023scaffold,cheng2024gaussianpro,ren2024octree},
% compression and regularization \cite{navaneet2023compact3d,lee2023compact,girish2023eagles,niedermayr2023compressed},
% physical simulation \cite{wu20234d,liang2023gaufre,duan20244d,sun20243dgstream} 
% and so on.

% For the geometry reconstruction,
% NeuSG~\cite{chen2023neusg} incorporate the previous NeRF-based surface reconstruction method NeuS in the 3DGS representation
% to transfer the surface property to 3DGS.
% SuGaR~\cite{guedon2023sugar} extracts explicit meshes from the 3DGS representation by regularizing Gaussians over surfaces. 
% Based on SuGaR, GaMeS~\cite{waczynska2024games} use the explicit mesh as input and parameterizes Gaussian components using the vertices, which can modify Gaussians in real time by altering mesh components during inference. PGSR used a planer-based gaussian representation and purpose a unbiased depth rendering method. However, these methods are mainly designed for object-centric camera trajectory and are not fit for urban street scenes which is long and narrow. We make the first attempt to conduct scalable urban scene surface reconstruction.

% Scaffold-GS~\cite{lu2023scaffold}
% Octree-GS~\cite{ren2024octree}

%第一个街景重建

\section{Preliminaries on 3D-GS and Octree Representation}
% 3D Gaussian Splatting~\cite{kerbl20233d} explicitly models the scene using anisotropic 3D Gaussian primitives and generates images via rasterizing projected 2D Gaussians, achieving SOTA rendering performance in novel view synthesis task within neural scene representations.
% Each 3D Gaussian $G(x)$ is parameterized by a center position $\mu \in \mathbb{R}^3$  and the covariance $\Sigma \in \mathbb{R}^{3 \times 3}$:
The method of 3D Gaussian Splatting~\cite{kerbl20233d} represents scenes by using stretched 3D Gaussian blobs and creates images by flattening them onto a 2D plane. This approach has achieved top-level performance in tasks involving novel view generation and neural scene representations. Each Gaussian in 3D space, denoted by $\mathbf{H}(\mathbf{x})$, is defined by a central point $\bm{\mu} \in \mathbb{R}^3$ and a covariance matrix $\bm{\Sigma} \in \mathbb{R}^{3 \times 3}$, and is described mathematically by the following expression:
\begin{equation}
\mathbf{H}(\mathbf{x}) = \exp\left(-\frac{1}{2} (\mathbf{x} - \bm{\mu})^T \bm{\Sigma}^{-1} (\mathbf{x} - \bm{\mu})\right),
\end{equation}
where $\mathbf{x}$ represents a point within the scene, and $\bm{\Sigma}$ is the covariance matrix, which can be factorized as $\bm\Sigma = \mathbf{R} \mathbf{S} \mathbf{S}^T\mathbf{R}^T$, with $\mathbf{R}$ being a rotation matrix and $\mathbf{S}$ a diagonal matrix indicating the scaling.
% where $x$ is an arbitrary position within the scene,
% %
% $\Sigma$ is parameterized by a scaling matrix $S \in \mathbb{R}^3$ and rotation matrix $R \in \mathbb{R}^{3 \times 3}$ with $R S S^T R^T$.
% For rendering, opacity $\sigma \in \mathbb{R}$ and color feature $F \in \mathbb{R}^C$ are represented using spherical harmonics (SH) to model view-dependent color $c \in \mathbb{R}^3$. The 3D Gaussians $G(x)$ are projected onto the image plane as 2D Gaussians $G^{\prime}(x^{\prime})$ \cite{zwicker2001ewa}. 
% A tile-based rasterizer efficiently sorts them in front-to-back depth order and employs $\alpha$-blending:
An optimized rasterization pipeline sorts these 2D Gaussians by depth and applies a blending technique based on their transparency. The color at each pixel on the 2D plane is calculated through:

\begin{equation}
    \mathbf{C}=\sum_{i=1 }^n{T_i \alpha_i\mathbf{c}_i }, 
    %\quad \alpha_i=G_i^{\prime}\left(x^{\prime}\right),
    \quad
    T_i = \prod_{j=1}^{i-1}\left(1-\alpha_j\right)
     % \mathbf{C}(x') = \sum_{k=1}^{m} \kappa_k \mathbf{g}_k \lambda_k, \quad \lambda_k = \gamma_k H'_k(x'),
\end{equation}
where $n$ represents the number of 2D Gaussians contributing to that pixel, $T$ refers to the transmission factor, $\mathbf{c}_i$ is the spherical harmonics representation of the $i$-th Gaussian.

% where $x^{\prime}$ is the queried pixel, $n$ represents the number of sorted 2D Gaussians binded with that pixel, and $T$ denotes the transmittance as $\prod_{j=1}^{i-1}\left(1-\sigma_j\right)$.

Subsequent Scaffold-GS~\cite{lu2023scaffold} uses anchor points for 3D scene gaussian representation. Builing on this, Octree-GS~\cite{ren2024octree} aligns these anchor points with level-of-detail (LOD) requirements to form a multi-level representation of the scene, which enhances rendering quality and increases rendering speed.

%scaffold,octree

% Subsequent PGSR~\cite{chen2024pgsr} introduces a planer-based Gaussian splatting representation that achieves high-fidelity surface reconstruction while ensuring high-quality rendering. It introduces an unbiased depth rendering method, which directly renders the distance from the camera origin to the Gaussian plane and the corresponding normal map based on the Gaussian distribution of the
% point cloud, and divides the two to obtain the unbiased depth.

\section{Methods}
We address the challenge of reconstructing and rendering high-fidelity geometry of static urban scenes using multi-view RGB images. Our methodology initiates with constructing an octree representation based on anchor Gaussians derived from a sparse Structure-from-Motion (SfM) point cloud. These Gaussians are dynamically assigned across different Levels of Detail (LODs) and progressively flattened into 2D Gaussian representations. Using this flattened Gaussian model, we generate plane distance and normal maps, subsequently transforming these into depth maps. 
%To deal with objects overlappings and improve the robustness of our method, we implement outlier and edge filtering to improve the accuracy of geometry in depth and normal maps from single views, tailoring our approach to urban scene characteristics. 
We employ techniques such as outlier and edge filtering to enhance the accuracy of geometry in depth and normal maps, particularly in regions with object overlap, tailoring our approach to the characteristics of urban scenes. 
Additionally, we enhance multi-view consistency by integrating both neighboring camera matching and far-field matching strategies to assess photometric and geometric consistencies across views. Our training approach includes a progressive strategy to underscore the distinct roles of various LODs, alongside a segmented training strategy to enhance scalability. Following the training phase, we use TSDF Fusion to extract the final mesh structure.

% We also adaptively refine the anchor points to capture more details and remove outliers Gaussian points using a novel growing and pruning operation for Gaussian density control with a learnable LOD bias to adjust anchor details between adjacent levels. Initially, we improve the modeling of scene geometry attributes by compressing. 3D Gaussians into a 2D flat plane representation, which is used to generate plane distance and normal maps, and subsequently converted into depth maps. We then introduce single-view depth and normal geometry consistency constraints to ensure the correctness of single-view geometry. Next, we incorporate multi-view photometric and geometric consistency loss to ensure global geometry consistency. Additionally, we use an implicit exposure model to account for brightness variations caused by exposure, further improving outdoor reconstruction accuracy.
% Finally, we propose Gaussian truncation and addition strategies to reduce the number of Gaussians while maintaining scene reconstruction quality.

\subsection{Planar-based Octree Gaussian Splatting Representation}

% In this subsection, we present our novel planar-based octree Gaussian splatting representation designed for the complex task of urban scene reconstruction.
% To achieve high-quality surface construction while maintaining geometric accuracy, we adopt the planar-based representation as outlined in PGSR. By integrating this with an octree architecture, we effectively accommodate the elongated characteristics typical of street views, reducing memory consumption and enhancing rendering efficiency.

Our approach begins with a sparse point cloud generated using structure-from-motion (SfM) with COLMAP~\cite{schonberger2016structure}. Following the methodologies described in ~\cite{lu2023scaffold} and~\cite{ren2024octree}, we approximate an axis-aligned bounding box for the scene and partition the sparse point cloud via an octree-based scheme. Details of anchor initialization can be found in~\cite{ren2024octree}. 
% Each input camera view is associated with its corresponding anchor Gaussians, optimizing the octree structure's utility. The depth range bounding pairs $(i, j)$ are estimated as $(\hat{d}_{\text{min}}, \hat{d}_{\text{max}})$ using the $0.05$ and $0.95$ quantiles from the depth distribution $\mathbf{D} = \{d_{i,j}\}$, incorporating scaling factors to accommodate varying camera intrinsics. The number of octree layers $K$ is determined by:
% \begin{equation}
% K = \left\lfloor \log_{2} \left( \frac{\hat{d}_{\text{max}}}{\hat{d}_{\text{min}}} \right) \right\rfloor + 1.
% \end{equation}

% The centers of the occupied voxels $\mathbf{V} \in \mathbb{R}^{M \times 3}$ are computed as:
% \begin{equation}
% \mathbf{V} = \left\{ \left\lfloor \frac{\mathbf{P}}{\epsilon} \right\rceil \cdot \epsilon, \ldots, \left\lfloor \frac{\mathbf{P}}{\epsilon / 2^{K-1}} \right\rceil \cdot \epsilon / 2^{K-1} \right\},
% \end{equation}
% where the base voxel size $\epsilon$ corresponds to the coarsest octree layer. The level of detail (LOD) for each anchor Gaussian is recorded in $\mathbf{L} \in \mathbb{R}^{M \times 1}$, forming a multi-level representation of the scene.

After structuring the octree and identifying the voxel centers, we convert the 3D Gaussians into 2D flat Gaussians to more accurately capture the surface geometry of the scene. Using the planar-based octree Gaussian splatting representation, we render depth and normal maps for subsequent regularizations, ensuring a precise and efficient urban scene reconstruction.

\begin{figure}[tbp]
    \centering

        \includegraphics[width=\linewidth]{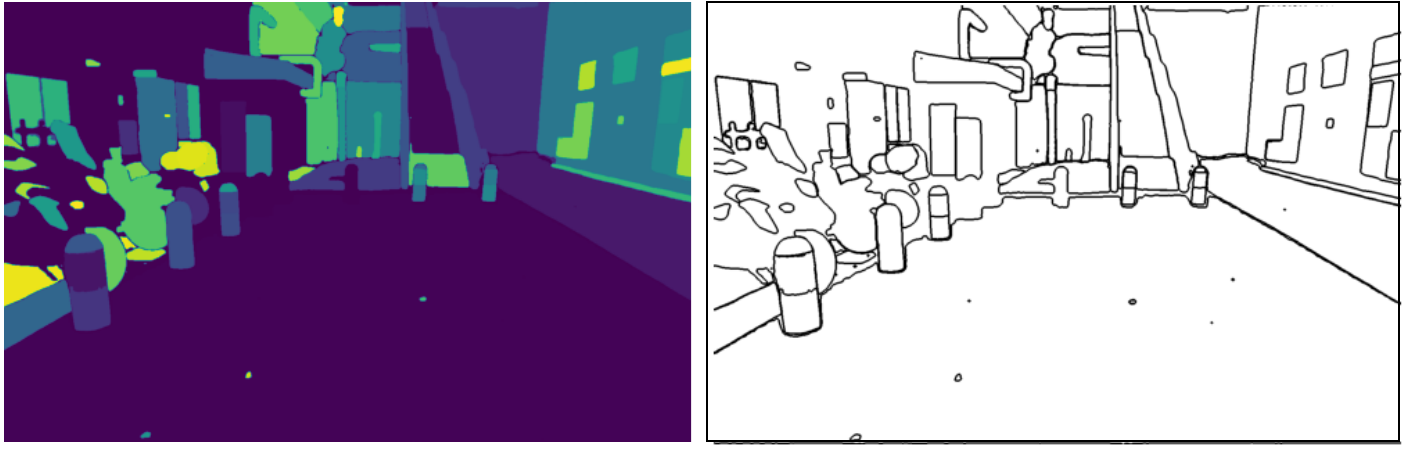}
        \caption{Illustration of our edge filtering strategy. We use SAM to generate a mask, followed by the extraction of width-defined boundaries for removal within the smoothness constraint.}
        \label{fig:method2}
\end{figure}
   \begin{figure}[tbp]
        \centering
        \includegraphics[width=\linewidth]{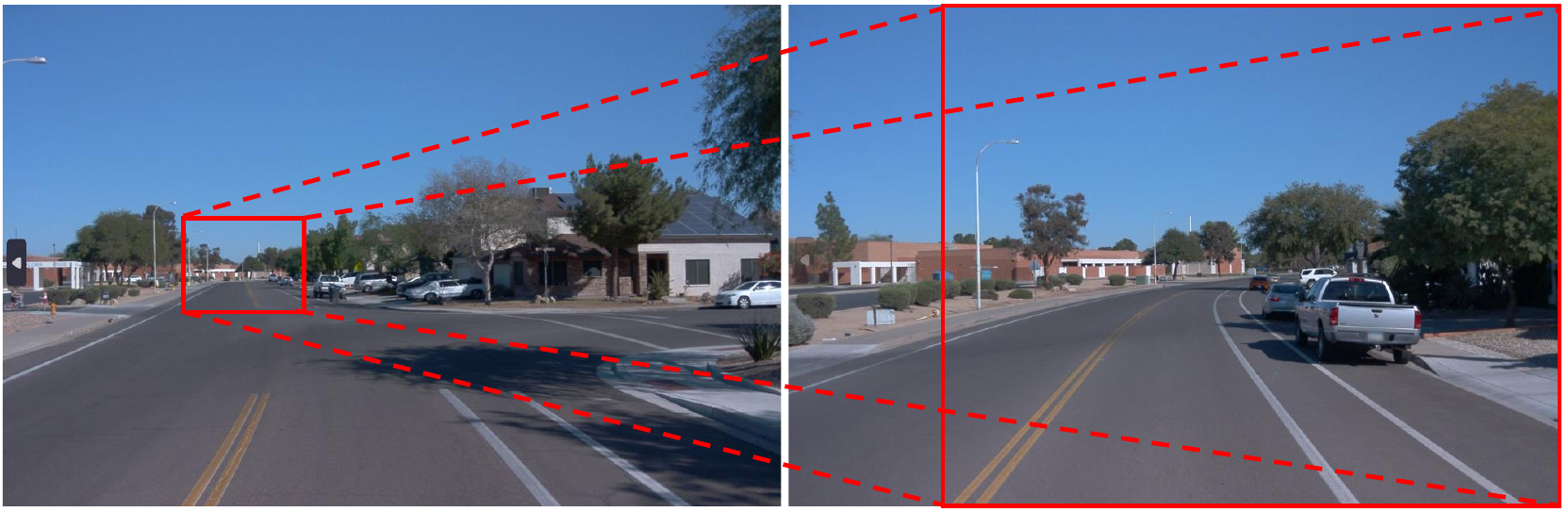}
        \caption{Illustration of our far-field matching strategy. We enforce long-term constraints by considering the consensus regions of distant frames to reduce accumulated error.}
        \label{fig:method1}
\end{figure}

\subsection{Regularization}
With depth and normal maps
available, we conduct specially designed smooth regularization and multi-view consistency regularization to improve the accuracy and consistency of geometry information.
% With geometric depth, distance and normal maps
% available, we conduct specially designed smooth regularization and multi-view consistency regularization to improve the accuracy and consistency of geometry information.

\subsubsection{Smooth Regularization}

To ensure a consistent alignment between the normals derived from the rendered depth map and the normals calculated directly from the Gaussian planes, we enforce a smooth regularization based on neighborhood analysis. Specifically, for each point in the rendered depth map, we calculate the normal vector using the \textbf{neighboring four points}: 
\begin{equation}
    \mathbf{N}_d(\mathbf{p}) = \frac{(\mathbf{P}_2(\mathbf{p}) - \mathbf{P}_1(\mathbf{p})) \times (\mathbf{P}_4(\mathbf{p}) - \mathbf{P}_3(\mathbf{p}))}{\|(\mathbf{P}_2(\mathbf{p}) - \mathbf{P}_1(\mathbf{p})) \times (\mathbf{P}_4(\mathbf{p}) - \mathbf{P}_3(\mathbf{p}))\|},
\end{equation}
where $\mathbf{P}_1(\mathbf{p})$ to $\mathbf{P}_4(\mathbf{p})$ correspond to the spatial positions of the four neighbors of point $\mathbf{p}$ in the depth map.
We then impose consistency between these computed normals and those directly derived from the Gaussian representation:
\begin{equation}
\mathcal{L}_{\text{normal}} = \| M(\mathbf{p}) \cdot (\mathbf{N}_d(\mathbf{p}) - \mathbf{N}(\mathbf{p})) \|_1,
\end{equation}
where $\mathbf{N}$ is the rendered normal map, $M(\mathbf{p})$ acts as a binary mask, where its value is 1 for valid points and 0 for invalid points. 
To address the irregularities in the shape and layout of objects in street scenes,  we employ outlier filtering and edge filtering to construct our mask.

%address weak texture regions, and mitigate the effects of varying lighting conditions,

\textbf{Outlier Filtering.} 
For the outlier filtering mask $M_1(\mathbf{p})$, we calculate the angle between $\mathbf{N}(\mathbf{p})$ and $\mathbf{N}_d(\mathbf{p})$. Points with excessively large angles are considered outliers, as they are likely not part of flat regions and should not be subjected to smoothness constraints. 

\textbf{Edge Filtering.}
In edge filtering, we adopt a two-step approach to get the edge filtering mask $M_2(\mathbf{p})$. As shown in Figure \ref{fig:method2}, a segmentation model, such as the Segment Anything Model (SAM) is first employed to generate a mask where each pixel is assigned a value representing its corresponding class or region within the image. Subsequently, we refine this mask by identifying boundaries between different regions. By filtering out these edge points which are near the boundaries, we further enhance the accuracy of the normal consistency.

The ultimate filtering mask $M(\mathbf{p})$ is the combination of these two filtering masks: 
\begin{equation}
    M(\mathbf{p})= M_1(\mathbf{p}) \cdot M_2(\mathbf{p}).
\end{equation}

% \textbf{Outlier Filtering.}
% To eliminate outlier points where the computed normal significantly deviates from the rendered normal, we filter out any point pairs where the angle between the two normals exceeds a threshold, thus considering them as outliers.

% we identify edge points using a segmentation model, such as the Segment Anything Model (SAM), which helps identify and remove points located on object boundaries. By filtering out these edge points, we further enhance the accuracy of the normal consistency, resulting in smoother, more reliable Gaussian surface representations.

%\end{wraptable}
% \end{minipage}
%  \begin{minipage}[b]{0.49\linewidth}
% \begin{table}[htbp]

\subsubsection{Multi-view Consistency Regularization}
% To address the challenge of sparse viewpoints inherent in street view scenes and the need to simultaneously handle close-up details and distant backgrounds in street view reconstruction, we employ a dual strategy for multi-view consistency regularization. On one hand, we incorporate additional constraints from more adjacent viewpoints to strengthen local geometric and photometric consistency. To integrate extensive street view data, ensure spatial consistency and coherence in the reconstruction results, reduce the impact of accumulated errors, and fill in information gaps in sparse regions, we leverage longer-term information and establish far-field matching constraints. This approach capitalizes on the common areas across distant frames to ensure robust and coherent 3D reconstructions despite the sparse and varied nature of street view data.

In addressing sparse viewpoints inherent in street view scenes and the challenge of handling close-up details and distant backgrounds simultaneously, we adopt a dual strategy for multi-view consistency regularization, leveraging both additional constraints from adjacent viewpoints and far-field matching constraints to ensure robust and coherent 3D reconstructions.

% To address the challenge of sparse viewpoints inherent in street view scenes, we employ a dual strategy for multi-view consistency regularization. On one hand, we incorporate additional constraints from more adjacent viewpoints to strengthen local geometric and photometric consistency. On the other hand, unlike object-centric scenes which have many shared regions, each region in the street scenes has relatively sparse views. To leverage longer-term information, we use key frame strategy to choose frames and establish far-field matching constraints. This approach capitalizes on the common areas across distant frames to ensure robust and coherent 3D reconstructions despite the sparse and varied nature of street view data.

\textbf{Neighbour camera matching.}
We further expand the neighbour camera matching strategy used in PSGR~\cite{chen2024pgsr}.
For each pose, we select two cameras from the four nearest viewpoints. Depth maps are generated, and key points from the current viewpoint are transformed into the nearest cameras' coordinate systems. These points are projected back into the current view to measure pixel-wise differences. Additionally, we compute the normalized cross-correlation (NCC) between patches in the reference and nearest images to enforce photometric consistency. 

\textbf{Far-field matching.}
% As shown in Figure \ref{fig:method1}, in street view scenes, since the camera continuously moves forward while capturing images, many parts of the subsequent frames will be magnified versions of the previous frames. This provides an opportunity for us to implement far-field matching. far-field matching involves selecting pairs of cameras that are far apart and computing a geometric consistency loss based on their depth maps.
As illustrated in Figure \ref{fig:method1}, in street view scenes, the continuous forward movement of the camera while capturing images results in many regions of subsequent frames being magnified versions of those in earlier frames. This characteristic facilitates the implementation of far-field matching. Specifically, we exploits this magnification phenomenon by selecting pairs of cameras that are significantly separated. By computing a geometric consistency loss based on their respective depth maps, we %leverage long-term information and 
establish correspondences across distant viewpoints.

For each given pose, 
we employ a key frame strategy to identify distant cameras, ensuring significant angular differences or minimum distances between selected viewpoints. From these distant cameras, we randomly select two for matching.
The following operations are conducted on one of the selected far cameras, with the other undergoing the same procedure. 
Depth maps \(\mathbf{D}_1\) and \(\mathbf{D}_2\) are generated for both the given pose and the selected far camera.  The relative transformation, composed of rotation matrix \(\mathbf{R}\) and translation vector \(\mathbf{t}\), between the viewpoint and far camera is calculated as follows:
\begin{equation}
    \mathbf{R} = \mathbf{R}_2 \mathbf{R}_1^\top, \quad \mathbf{t} = \mathbf{t}_2 - \mathbf{R} \mathbf{t}_1,
\end{equation}
where \(\mathbf{R}_1\) and \(\mathbf{R}_2\) represent the rotation matrices of the viewpoint and the far camera, respectively, and \(\mathbf{t}_1\) and \(\mathbf{t}_2\) denote the translation vectors of the viewpoint and the far camera, respectively.

This transformation maps the far camera's depth values into the viewpoint camera's coordinate system. The far camera's depth map is then resampled onto the viewpoint camera's grid. 
Relative depth errors between these two depth maps are computed using the equation:
%Relative depth errors between the transformed far camera's depth map and the viewpoint camera's depth map are computed using this equation:
\begin{equation}
    \Delta \mathbf{D} = \left| \frac{\mathbf{D}_1 - \mathbf{D}_2'}{\mathbf{D}_1 + \epsilon} \right|,
\end{equation}
where \(\mathbf{D}_2'\) is the transformed depth map of the far camera, and \(\epsilon\) is a small constant to avoid division by zero. This equation denotes point-wise operations on the depth map values. These errors are averaged over valid, non-zero, and co-visible regions to obtain the final geometric consistency loss.

%we identify 
%distant cameras using a key frame strategy and randomly select two of them. The key frame strategy ensures that the chosen viewpoints have a significant angular difference or are separated by a minimum distance.

%For each given pose, we identify %cameras that are sufficiently distant from the current viewpoint 
% distant cameras using a key frame strategy and randomly select two of them. The key frame strategy ensures that the chosen viewpoints have a significant angular difference or are separated by a minimum distance.
%thereby capturing diverse perspectives. 

% As illustrated in Figure \ref{fig:method1}, in street view scenes, the continuous forward movement of the camera while capturing images results in many regions of subsequent frames being magnified versions of those in earlier frames. This characteristic facilitates the implementation of far-field matching. far-field matching involves selecting pairs of cameras that are significantly separated and computing a geometric consistency loss based on their respective depth maps.

% These errors are averaged over valid, non-zero, and co-visible regions to obtain the final geometric consistency loss. By enforcing this loss, we achieve more accurate 3D reconstructions and improve the overall quality of the generated models, as the far-field matching process robustly ensures consistency across widely varying viewpoints. 
%This technique is a crucial component of our method, significantly contributing to the robustness and accuracy of the 3D reconstruction results.

\subsection{Training Strategy.}
In our setting, where the data stream is continuously evolving, early-stage information holds limited significance for subsequent modeling. To ensure scalability and enhance local modeling fidelity in our approach, we employ a segmented training strategy. Concurrently, to bolster the stability of training across different LODs, we implement progressive training strategy.

\textbf{Segmented Training.}
To handle the dynamic nature of the data stream, our training strategy segments the captured video frames temporally with intentional overlap between segments. This design allows each segment to be modeled individually while leveraging continuity in the video stream. Specifically, the overlapping portions of the segments use depth information rendered from the preceding segment to guide the training of the following segment, ensuring a seamless transition and temporal coherence. 
% while enhancing depth prediction accuracy.

\textbf{Progressive Training.}
% To mitigate inadequate supervision at higher levels, we refrain from optimizing anchor Gaussians simultaneously across all LOD levels. Drawing from prior progressive training strategies~\cite{park2021nerfies, xiangli2022bungeenerf, li2023neuralangelo}, we employ a coarse-to-fine optimization approach. Beginning training at lower LOD levels and gradually activating higher levels, this method effectively organizes anchor points across the LOD spectrum.
We avoid optimizing anchor Gaussians simultaneously across all LOD levels to prevent insufficient supervision at higher levels. %Inspired by previous progressive training strategies~\cite{park2021nerfies, xiangli2022bungeenerf, li2023neuralangelo}, 
Specifically, we implement a coarse-to-fine optimization approach, starting training at lower LOD levels and progressively activating higher levels, so that we can organizes anchor points across the LOD spectrum. 

%This method effectively organizes anchor points across the LOD spectrum, reducing per-view Gaussian counts for faster rendering while preserving high-quality outputs.

% Optimizing anchor Gaussians across all LOD levels simultaneously can lead to insufficient supervision at higher levels, often missing fine-grained details. Inspired by previous progressive training strategies~\cite{park2021nerfies, xiangli2022bungeenerf, li2023neuralangelo}, we adopt a coarse-to-fine optimization. Training starts at the lower LOD levels, progressively activating higher levels. This approach organizes anchor points more effectively across the LOD spectrum, reducing per-view Gaussian counts and enabling faster rendering while maintaining high-quality outputs.

\textbf{TSDF Fusion.}
We concatenate the rendered images and depths obtained from each segment in the segmented training and use the TSDF Fusion algorithm to generate the corresponding TSDF field. 
Subsequently, we extract a mesh from the TSDF field and apply mesh cleaning to remove noise, resulting in the final mesh. Hyperparameter discussion is shown in the appendix.

\begin{table}[tbp]
    \centering
    \small
    \caption{Performance comparison for novel view synthesis on the KITTI-360 dataset~\cite{liao2022kitti}.  `*' indicates results evaluated from trained models using the official implementation. All other results are sourced from DNMP~\cite{lu2023urban}.}
    \label{tab:kitti}
    \begin{tabular}{@{}l@{\hskip 3pt}@{\hskip 3pt}c@{\hskip 8pt}c@{\hskip 8pt}c@{}}
        \toprule
        \textbf{Method} & PSNR{\scriptsize$\uparrow$} & SSIM{\scriptsize$\uparrow$} & LPIPS{\scriptsize$\downarrow$} \\
        \midrule
        \multicolumn{4}{c}{\textbf{No mesh (except Ours)} } \\
        \midrule
        NeRF~\cite{mildenhall2021nerf} & 21.94 & 0.781 & 0.449 \\
        NeRF-W~\cite{martin2021nerf} & 22.77 & 0.794 & 0.446 \\
        Instant-NGP~\cite{mueller2022instant} & 22.89 & 0.836 & 0.353 \\
        Point-NeRF~\cite{xu2022point} & 21.54 & 0.793 & 0.406 \\
        BungeeNeRF~\cite{xiangli2022bungeenerf} & 22.02 & 0.788 & 0.429 \\
        Mega-NeRF~\cite{turki2022mega} & 23.15 & 0.826 & 0.326 \\
        Mip-NeRF~\cite{barron2021mip} & 23.21 & 0.810 & 0.455 \\
        Mip-NeRF 360~\cite{barron2022mip} & 23.27 & 0.836 & 0.355 \\
        3DGS*~\cite{kerbl20233d} &22.96&0.846& 0.251\\
        Ours &\textbf{23.90} &\textbf{0.874} &\textbf{0.246} \\
        \midrule
        \multicolumn{4}{c}{\textbf{With mesh} } \\
        \midrule
        MVS~\cite{metashape} & 21.82 & 0.817 & 0.341 \\
        Neus*~\cite{wang2021neus} & 20.72 & 0.781 & 0.333 \\
        Neuralangelo*~\cite{li2023neuralangelo} & 19.86 & 0.735 & 0.258 \\
        DNMP~\cite{lu2023urban} & 23.41 & 0.846 & 0.305 \\
        SuGaR*~\cite{guedon2023sugar} & 22.67 & 0.821 & 0.252 \\
        PGSR*~\cite{chen2024pgsr} &22.86&0.854&0.260 \\
        Ours &\textbf{23.90} &\textbf{0.874} &\textbf{0.246} \\
        \bottomrule
    \end{tabular}
\end{table}

\section{Experiment}

\begin{table}[htbp]
    \centering
    \small
    \caption{Performance comparison for novel view synthesis on the Waymo Open dataset~\cite{sun2020scalability}. `*' indicates results evaluated from trained models using the official implementation. All other results are sourced from DNMP~\cite{lu2023urban}.}
    \label{tab:waymo}
    \begin{tabular}{l c c c}
        \toprule
        \textbf{Method} & PSNR$\uparrow$ & SSIM$\uparrow$ & LPIPS$\downarrow$ \\
        \midrule
        \multicolumn{4}{c}{\textbf{No mesh (except Ours)} } \\
        \midrule
        NeRF~\cite{mildenhall2021nerf} & 26.24 & 0.870 & 0.472 \\
        NeRF-W~\cite{martin2021nerf} & 26.92 & 0.885 & 0.418 \\
        Instant-NGP~\cite{mueller2022instant} & 26.77 & 0.887 & 0.401 \\
        Point-NeRF~\cite{xu2022point} & 26.26 & 0.868 & 0.450 \\
        NPLF~\cite{ost2022neural} & 25.62 & 0.879 & 0.450 \\
        Mip-NeRF~\cite{barron2021mip} & 26.96 & 0.880 & 0.451 \\
        Mip-NeRF 360~\cite{barron2022mip} & 27.43 & 0.893 & 0.394 \\
        Ours &\textbf{28.67} &\textbf{0.935} &\textbf{0.279} \\
        \midrule
        \multicolumn{4}{c}{\textbf{With mesh} } \\
        \midrule
        DNMP~\cite{lu2023urban} & 27.62 & 0.892 & 0.381 \\
        SuGaR*~\cite{guedon2023sugar} & 26.84 & 0.876 & 0.316 \\
        Ours &\textbf{28.67} &\textbf{0.935} &\textbf{0.279} \\
        \bottomrule
    \end{tabular}
\end{table}

\begin{table}[htbp]
    \centering
    \small
    \caption{Performance comparison for novel view synthesis on the Free dataset~\cite{wang2023f2}. `*' indicates results evaluated from trained models using the official implementation. All other results are sourced from F2NeRF~\cite{wang2023f2}.}
    \label{tab:compare_free}
    \begin{tabular}{@{}l@{\hskip 3pt}@{\hskip 3pt}c@{\hskip 8pt}c@{\hskip 8pt}c@{}}
        \toprule
        \textbf{Method} & PSNR{\scriptsize$\uparrow$} & SSIM{\scriptsize$\uparrow$} & LPIPS{\scriptsize$\downarrow$} \\
        \midrule
        \multicolumn{4}{c}{\textbf{No mesh (except Ours)} } \\
        \midrule
        NeRF++~\cite{ZhangRSK20} & 23.47 & 0.603 & 0.499 \\
        Plenoxels~\cite{YuFTCR22} & 19.13 & 0.507 & 0.543 \\
        DVGO~\cite{SunSC22} & 23.90 & 0.651 & 0.455 \\
        Instant-NGP~\cite{mueller2022instant} & 24.43 & 0.677 & 0.413 \\
        F2-NeRF~\cite{wang2023f2} & 26.32 & 0.779 & 0.276 \\
        Mip-NeRF-360~\cite{barron2022mip} & {\bf 27.01} & 0.766 & 0.295 \\
        Ours & 26.84 & {\bf 0.832} & {\bf 0.191} \\
        \midrule
        \multicolumn{4}{c}{\textbf{With mesh} } \\
        \midrule
        SuGaR*~\cite{guedon2023sugar} & 24.62 & 0.730 & 0.304 \\
        Ours & {\bf 26.84} & {\bf 0.832} & {\bf 0.191} \\
        \bottomrule
    \end{tabular}
\end{table}

\begin{table*}[htb]
    \centering
    \small
        \caption{Scene breakdown on the Free dataset~\cite{wang2023f2}. '-' means CUDA OUT OT MEOMORY.}
        \vspace{-6pt}
    \label{tab:compare_free_break_down}
    \begin{tabular}{lccccccc}
    \toprule
    Method & Hydrant & Lab & Pillar & Road & Sky & Stair & Grass \\
    \midrule
    \multicolumn{8}{c}{\textbf{No mesh} } \\
    \midrule
    NeRF++~\cite{ZhangRSK20} & 22.21 & 21.82 & 25.73 & 23.29 & 23.91 & 26.08 & 21.26 \\

    Plenoxels~\cite{YuFTCR22} & 19.82 & 18.12 & 18.74 & 21.31 & 18.22 & 21.41 & 16.28 \\
    DVGO~\cite{SunSC22} & 22.10 & 23.78 & 26.22 & 23.53 & 24.26 & 26.65 & 20.75 \\
    Instant-NGP~\cite{mueller2022instant} & 22.30 & 23.21 & 25.88 & 24.24 & 25.80 & 27.79 & 21.82 \\
     F2-NeRF~\cite{wang2023f2}&  24.34 &  25.92 &  28.76 &  26.76 &  26.41 &  29.19 &  22.87 \\
        Mip-NeRF 360~\cite{barron2022mip}  & 25.03 & 26.57 &  29.22&  27.07 & {\bf 26.99} & {\bf 29.79} & 24.39 \\
         3DGS~\cite{kerbl20233d} &-&26.70&28.47&-&26.49&28.70& -\\
         Ours &{\bf25.12}&{\bf26.89}&{\bf29.41}&{\bf27.18}&26.59&28.14& {\bf24.54}\\
         \midrule
             \multicolumn{8}{c}{\textbf{With mesh} } \\
    \midrule
         SuGaR~\cite{guedon2023sugar} &22.49&25.23&26.84&25.06&24.55&26.88& 21.25\\
         PGSR~\cite{chen2024pgsr} &-&26.57&28.05&-&-&27.84& -\\
         Ours &{\bf25.12}&{\bf26.89}&{\bf29.41}&{\bf27.18}&{\bf26.59}&{\bf28.14}& {\bf24.54}\\
    \bottomrule
    \end{tabular}

    \vspace{-0em}
\end{table*}

\begin{table}[htbp]
    \centering
    \small
    \caption{Memory usage on different scenes on the Free dataset~\cite{wang2023f2}.}
    \label{memory}
    \begin{tabular}{l c c c}
        \toprule
        \textbf{Method} & Stair & Pillar & Lab \\
        \midrule
        3DGS~\cite{kerbl20233d} & 17.93 GB
  & 14.29 GB
 & 17.12 GB  \\
        PGSR~\cite{chen2024pgsr} &23.26 GB
  &22.98 GB
 &23.65 GB \\
        Ours & 11.92 GB
  & 11.19 GB
 & 12.19 GB   \\
        \bottomrule
    \end{tabular}
\end{table}

\begin{table}[htbp]
    \centering
    \small
    \caption{Ablation studies conducted on the `Pillar' scene of the Free dataset~\cite{wang2023f2}. We start from a baseline model without these elements and then separately add or subtract each component.}
    \label{ablation}
    \begin{tabular}{l c c c}
        \toprule
        \textbf{Method} & PSNR{\scriptsize$\uparrow$} & SSIM{\scriptsize$\uparrow$} & LPIPS{\scriptsize$\downarrow$} \\
        \midrule
        baseline & 28.05 & 0.778 & 0.273 \\
        + octree & 28.45 & 0.823 & 0.241 \\
        + filtering & 28.47 & 0.820 & 0.228 \\
        + matching & \textbf{29.41} & \textbf{0.852} & \textbf{0.186} \\
        - filtering & 29.24 & 0.838 & 0.198 \\
        \bottomrule
    \end{tabular}
\end{table}

\begin{table}[htbp]
    \centering
    \small
    \caption{Ablation studies conducted on the first scene of the Kitti360 dataset~\cite{liao2022kitti}. We start from a baseline model without these elements and then separately add or subtract each component.}
    \label{ablation2}
    \begin{tabular}{l c c c}
        \toprule
        \textbf{Method} & PSNR{\scriptsize$\uparrow$} & SSIM{\scriptsize$\uparrow$} & LPIPS{\scriptsize$\downarrow$} \\
        \midrule
        baseline & 24.35 & 0.833 & 0.257\\
        + octree & 24.82 & 0.871 & 0.242 \\
        + filtering & 24.86 & 0.873 & 0.235 \\
        + matching & \textbf{25.90} & \textbf{0.914} & \textbf{0.226} \\
        - filtering & 25.63 & 0.887 & 0.229 \\
        \bottomrule
    \end{tabular}
\end{table}

\subsection{Experimental Setting}
\label{setting}
\noindent\textbf{Datasets.} 
We conduct experiments on four urban datasets: KITTI-360~\cite{liao2022kitti}, the Waymo Open dataset~\cite{sun2020scalability}, the Free dataset~\cite{wang2023f2}, and MatrixCity~\cite{li2023matrixcity}. 

% KITTI-360 is a large-scale dataset captured in urban environments with a driving distance of around $73.7$ km. 
% We follow \cite{lu2023urban} to select $5$ sequences from them to evaluate our method. 
% These sequences are relatively short, allowing us to compare StreetSurfGS with all the baseline methods, such as 3DGS~\cite{kerbl20233d} and PGSR~\cite{chen2024pgsr}.
% Following \cite{lu2023urban}, for the evaluation of novel view synthesis, we select every second image in each sequence as the test set and train our model on the remaining images. 
% For the evaluation of geometry reconstruction, we use all the images for training.

The KITTI-360 dataset, recorded across urban areas over a distance of approximately $73.7$ km, provides a large-scale dataset for our evaluation. Following the approach in \cite{lu2023urban}, we select $5$ specific sequences for testing our method. Due to the relatively short length of these sequences, we are able to directly compare the performance of StreetSurfGS against several baseline methods, including 3DGS~\cite{kerbl20233d} and PGSR~\cite{chen2024pgsr}. For novel view synthesis evaluation, we use every second frame as test data, training the model on the remaining images, as recommended by \cite{lu2023urban}. When assessing geometry reconstruction, all available frames are used for model training.

The Waymo Open dataset also represents urban scenes. Following the selection criteria in \cite{ost2022neural} and \cite{lu2023urban}, we use $6$ sequences that primarily capture static objects. Consistent with \cite{lu2023urban}, we designate every $10$th frame in these sequences as test data, while the remaining frames are used for training. The same experimental setup is applied to both novel view synthesis and geometry reconstruction tasks.

% Waymo Open is another urban environment dataset. We follow~\cite{ost2022neural} and \cite{lu2023urban} to select the $6$ sequences mainly containing static objects for our experiments. 
% Following \cite{lu2023urban}, we select every $10$th image in the sequences as the test set and take the remaining ones as the training set. We employ identical settings for both novel view synthesis and geometry reconstruction tasks.

% Free dataset contains
% seven scenes. Each scene has a narrow and long input camera
% trajectory and multiple focused foreground objects. It includes various urban elements like stairs, pillars, roads and grass which align well with the objectives of StreetSurfGS. Following \cite{wang2023f2}, we select every $8$th image in the sequences as the test set and take the remaining ones as the training set. We employ identical settings for both novel view synthesis and geometry reconstruction tasks. 

Free dataset contains
seven scenes. Each scene has a narrow and long input camera
trajectory and multiple focused foreground objects. It includes various urban elements like stairs, pillars, roads and grass which align well with the objectives of StreetSurfGS. Following \cite{wang2023f2}, we select every $8$th image in the sequences as the test set and take the remaining ones as the training set. We employ identical settings for both novel view synthesis and geometry reconstruction tasks.

MatrixCity~\cite{li2023matrixcity} is a comprehensive and
high-quality synthetic dataset to support the research of
city-scale neural rendering. We use all the images in the sequence for geometry evaluation. 

For images with resolutions exceeding 1600 pixels, we resize the longer side to 1600 pixels. %Dataset splits are provided in the appendix.

\noindent\textbf{Evaluation Metrics.} 
% Following the previous methods~\cite{mildenhall2021nerf,barron2021mip}, 
%our evaluations are based on three widely-used metrics, i.e., 
% We use peak signal-to-noise ratio (PSNR), structural similarity index measure (SSIM), and the learned perceptual image patch similarity (LPIPS)~\cite{zhang2018unreasonable} to evaluate rendering. For geometry, we use 
% mean squared error (MSE) to evaluate estimated normal vectors and depth map, Chamfer-Distance (C-D) and F1 Score to evaluate reconstructed point cloud.
%We employ a comprehensive suite of evaluation metrics for our rendering and geometry assessment. 
For rendering quality, we use peak signal-to-noise ratio (PSNR), structural similarity index measure (SSIM), and the learned perceptual image patch similarity (LPIPS)~\cite{zhang2018unreasonable}. Regarding geometric fidelity, we employ mean squared error (MSE) to assess the accuracy of estimated normal vectors and depth maps. Additionally, for evaluating the fidelity of reconstructed point clouds, we employ Chamfer-Distance (C-D) and F1 Score metrics.

\noindent\textbf{Baseline.}
\label{resource}
We compare our method against state-of-the-art reconstruction methods. For those focusing only on novel view synthesis, we include NeRF~\cite{mildenhall2021nerf},
NeRF-W~\cite{martin2021nerf}, 
Instant-NGP~\cite{mueller2022instant}, 
Point-NeRF~\cite{xu2022point}, 
BungeeNeRF~\cite{xiangli2022bungeenerf}, 
Mega-NeRF~\cite{turki2022mega}, 
NPLF~\cite{ost2022neural},
Mip-NeRF~\cite{barron2021mip}, 
Mip-NeRF 360~\cite{barron2022mip}
and F2-NeRF~\cite{wang2023f2}.
For methods incorporating mesh representations, we include MVS~\cite{metashape},
Neus~\cite{wang2021neus},
Neuralangelo~\cite{li2023neuralangelo},
DNMP~\cite{lu2023urban} and 
SuGaR~\cite{guedon2023sugar}. 
All baseline methods utilize Structure from Motion (SfM) points for initialization. To obtain surfaces for Neus~\cite{wang2021neus} and Neuralangelo~\cite{li2023neuralangelo}, we use a marching cube resolution of 2048, followed by referencing the Visibility and Free-space Culling from BakedSDF~\cite{yariv2023bakedsdf} to eliminate redundant meshes. 
To ensure fairness in our evaluation, we exclude StreetSurf~\cite{guo2023streetsurf} due to its reliance on monocular normals or LiDAR for loss supervision, which differs from the requirements of the other methods considered.
% To ensure fairness in our evaluation, we exclude StreetSurf~\cite{guo2023streetsurf} due to its reliance on monocular normals or LiDAR for loss supervision.
%external geometry information introduced through pretrained geometry estimation models.
All experiments in this paper are conducted on a single Nvidia RTX 3090 GPU. 

%Implementation details of our method are included in the appendix.

% \section{Results and Discussions}

\begin{figure*}[!ht]
\includegraphics[width=0.95\linewidth]{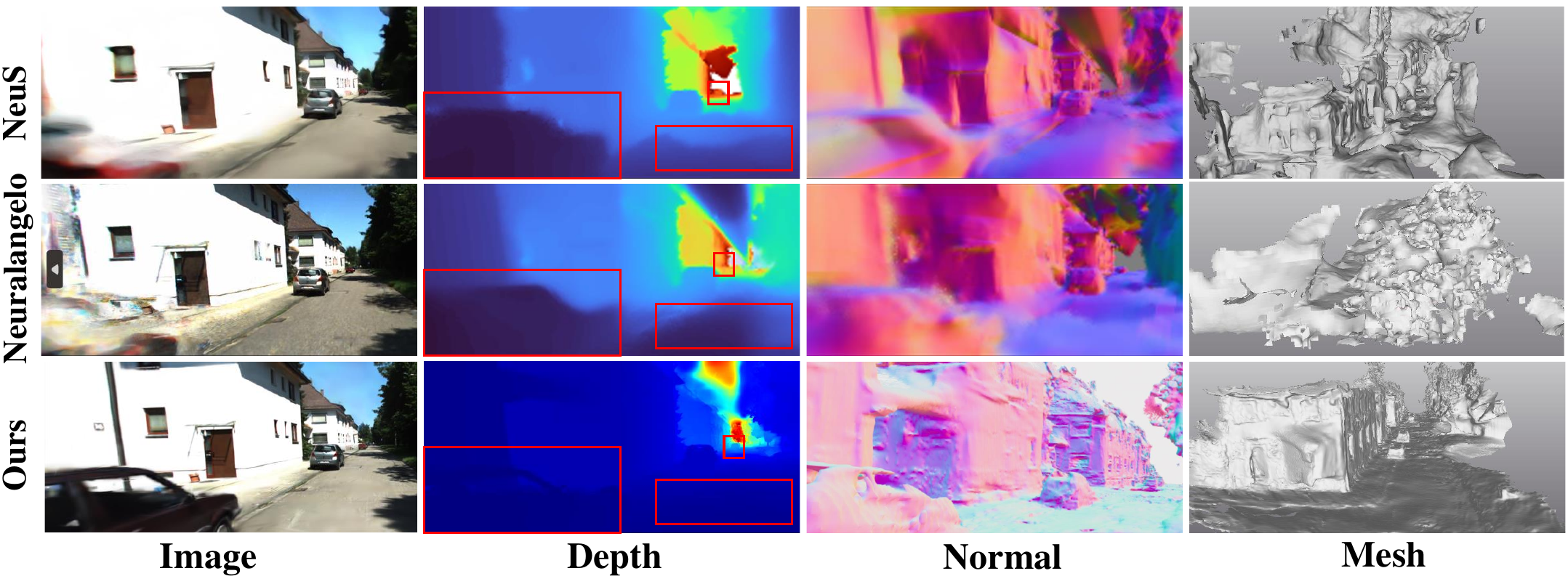}
\vspace{-5pt}
\caption{Overall Qualitative comparison on the KITTI-360~\cite{liao2022kitti}. Our filtering strategy enables accurate depth shifts, while our matching strategy effectively models objects at various distances. Additionally, our method constructs flat roads and generates mesh of significantly higher quality than previous methods.
}
\label{mesh_kitti}
\vspace{-5pt}
\end{figure*}

\begin{figure*}[!ht]
\centering
\includegraphics[width=0.95\linewidth]{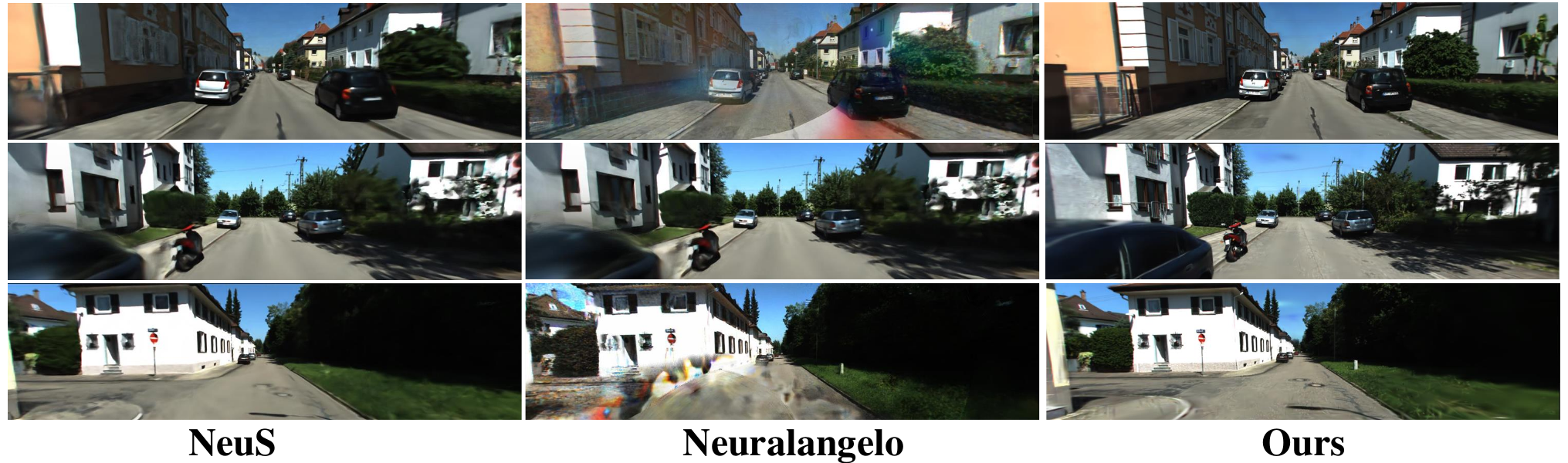}
\caption{Visual comparison for novel view synthesis on the KITTI-360~\cite{liao2022kitti}. Previous approaches predominantly depend on immediate frame data, which often results in blurred object peripheries. In contrast, our method extends the data scope by leveraging adjacent frame information coupled with long-term spatial constraints, which sharply reduces such blurring effects.
}
\label{v3}
\end{figure*}

\begin{figure*}[!ht]
\includegraphics[width=0.95\linewidth]{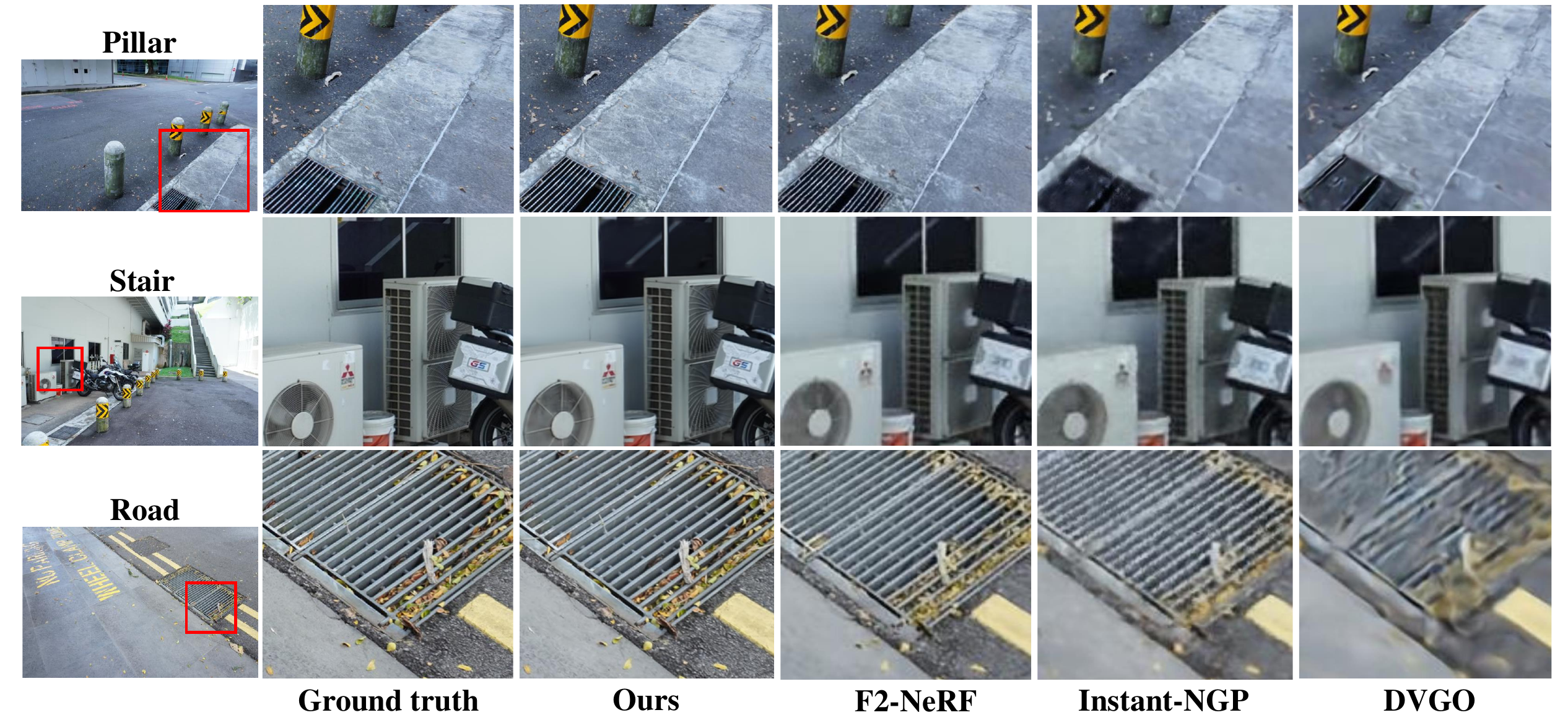}
\vspace{-5pt}
\caption{Visual comparison for novel view synthesis on the Free Dataset~\cite{wang2023f2}. Compared with images rendered by previous methods, our method produces clearer images with fine-grained details.
}
\label{render_free}
\end{figure*}

\begin{figure*}[!ht]
\centering
\includegraphics[width=0.95\linewidth]{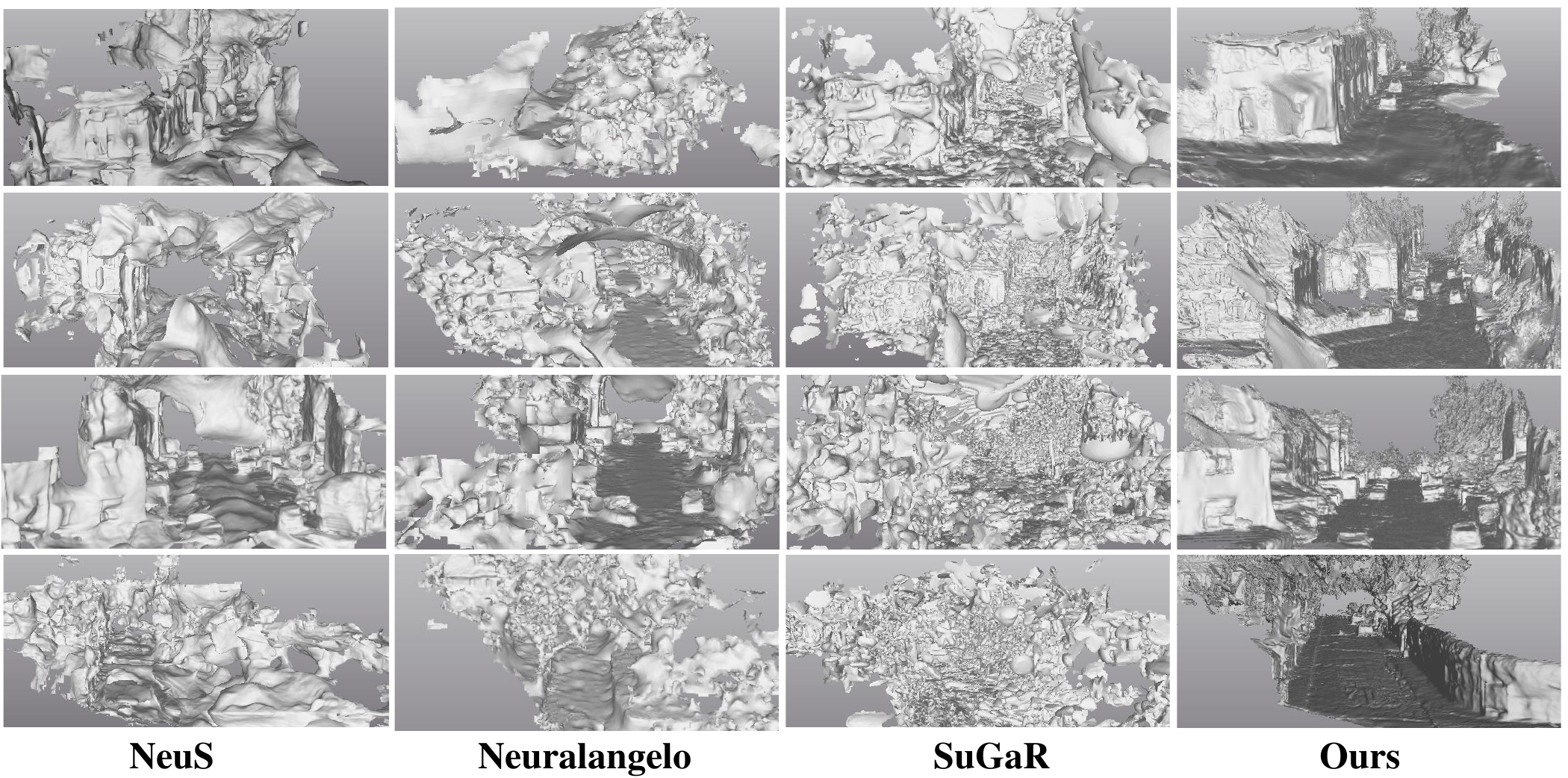}
\caption{
Qualitative comparison of surface reconstruction results on the KITTI-360~\cite{liao2022kitti}. Traditional techniques produce only rudimentary outlines and noisy mesh textures in urban street scenes. In contrast, our method not only delineates finer structural details but also ensures enhanced mesh fidelity.}
\label{v1}
\end{figure*}

\begin{figure*}[!ht]
\centering
\includegraphics[width=0.95\linewidth]{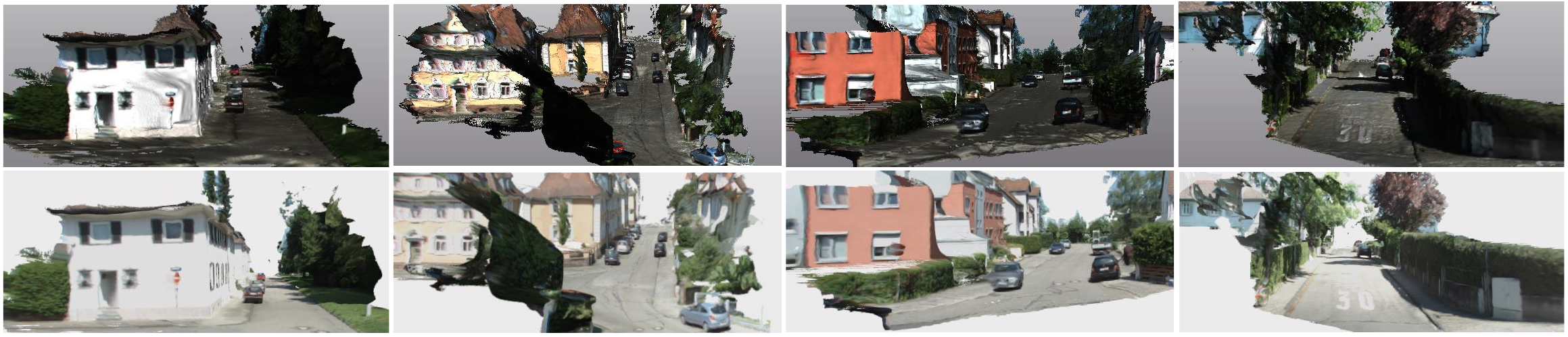}
\caption{Qualitative results of textured mesh. The first row is viewed using MeshLab, while the second row is viewed using the Open3D viewer.
}
\label{v2}
\end{figure*}

\begin{figure*}[ht]
    \centering
    \includegraphics[width=0.8\linewidth]{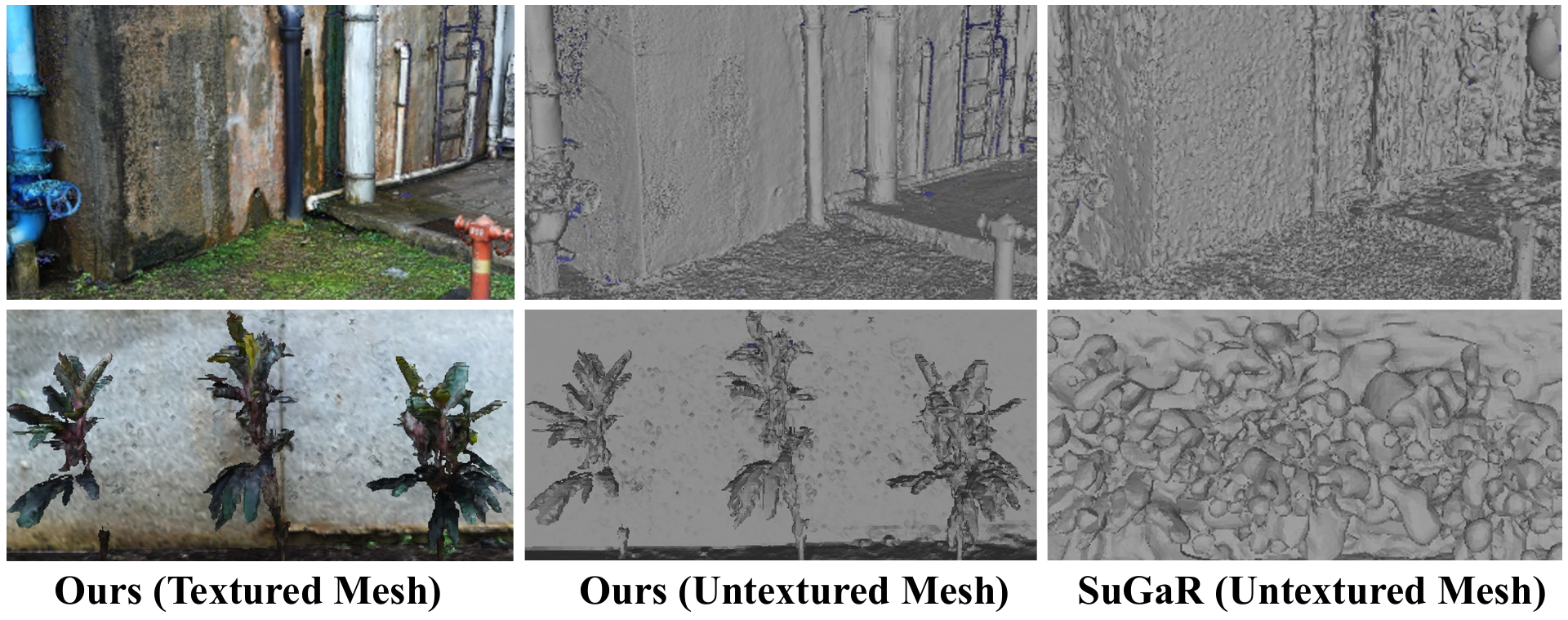}
    \caption{Visual comparison of surface reconstruction results on the Free dataset~\cite{wang2023f2}. Our technique has a better ability to create detailed and accurate meshes compared to existing methods.}
    \label{fig:mesh_free}
\end{figure*}

\begin{figure*}[!ht]
\centering
\includegraphics[width=0.95\linewidth]{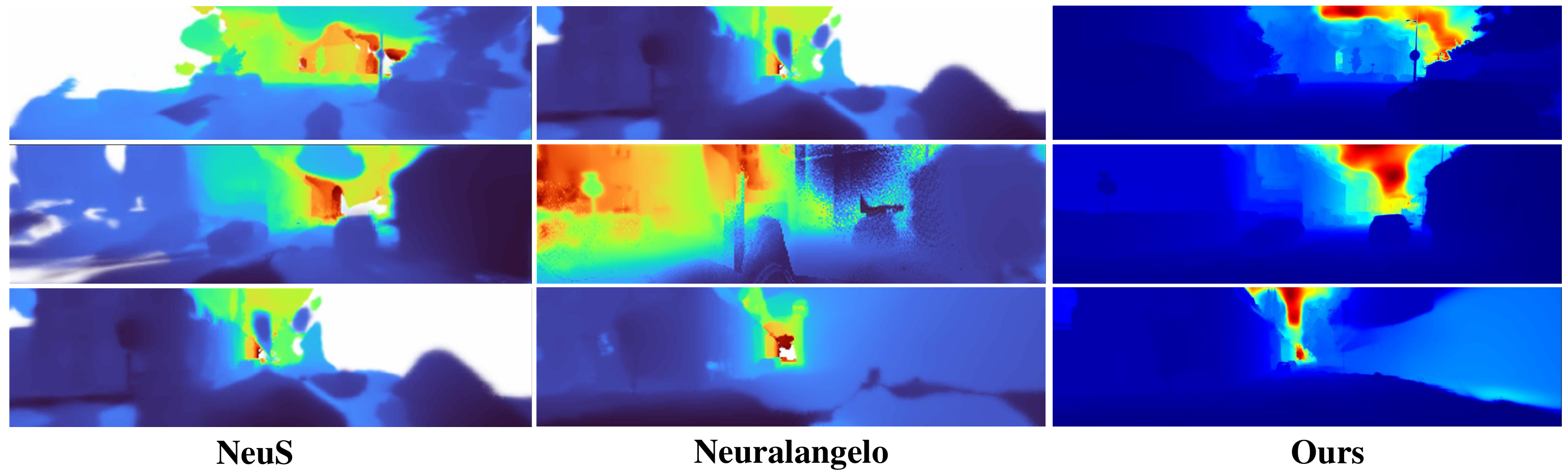}
\caption{Qualitative comparison of depth map on the KITTI-360~\cite{liao2022kitti}. Previous methods exhibit edge blurriness and poorly defined depth gradations. Our methodology addresses these deficiencies by incorporating long-term constraints that enhance far-field depth accuracy. 
}
\label{v4}
\end{figure*}

\begin{table}[tbp]
    \centering
    \caption{Performance comparison of geometry metrics on Matrixcity~\cite{li2023matrixcity}.}
    \label{matri}
    % \resizebox{0.5\linewidth}{!}{
    \begin{tabular}{lcc}
        \toprule
        \textbf{Method} & Depth Error$\downarrow$ & Normal Error$\downarrow$ \\
        \midrule
        Neus~\cite{wang2021neus} & 3.19 & 0.77 \\
        Neuralangelo~\cite{li2023neuralangelo} & 3.35 & 0.67 \\
        Ours & {\bf 1.51} & {\bf 0.42} \\
        \midrule
        \textbf{Method} & F1 Score$\uparrow$ & C-D$\downarrow$ \\
        \midrule
        Neus~\cite{wang2021neus} & 0.01 & 2.98 \\
        Neuralangelo~\cite{li2023neuralangelo} & 0.03 & 2.67 \\
        Ours & {\bf 0.13} & {\bf 0.29} \\
        \bottomrule
    \end{tabular}%}
    
\end{table}
% \end{wrapfigure}

\begin{figure*}[!ht]
\includegraphics[width=0.99\linewidth]{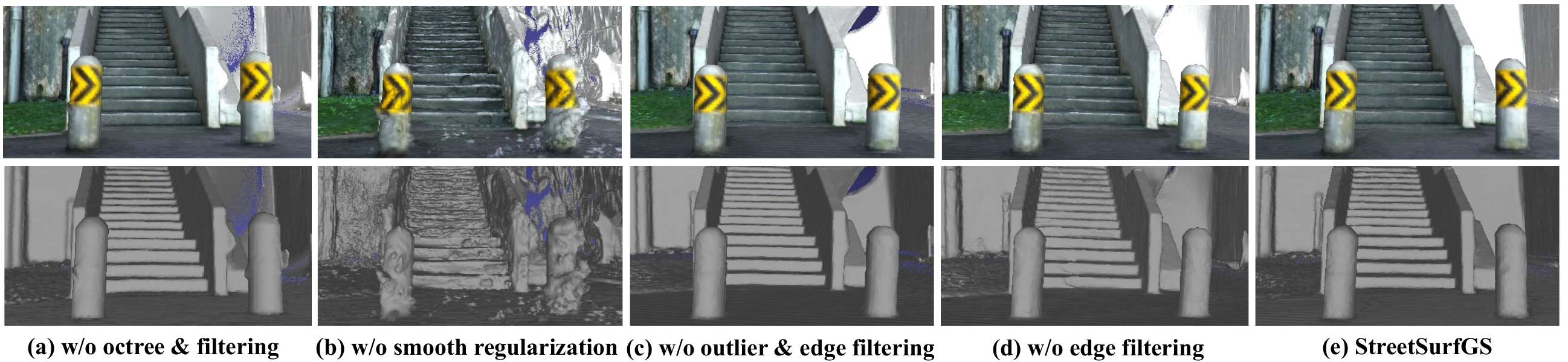}
\caption{Ablation study on reconstructed 
 mesh. The use of smooth regularization, along with outlier filtering and edge filtering, improves the accuracy of mesh reconstructions and provides a more natural smoothness.
}
\label{mesh_ablation}
\end{figure*}

\begin{figure*}[!ht]
\includegraphics[width=0.99\linewidth]{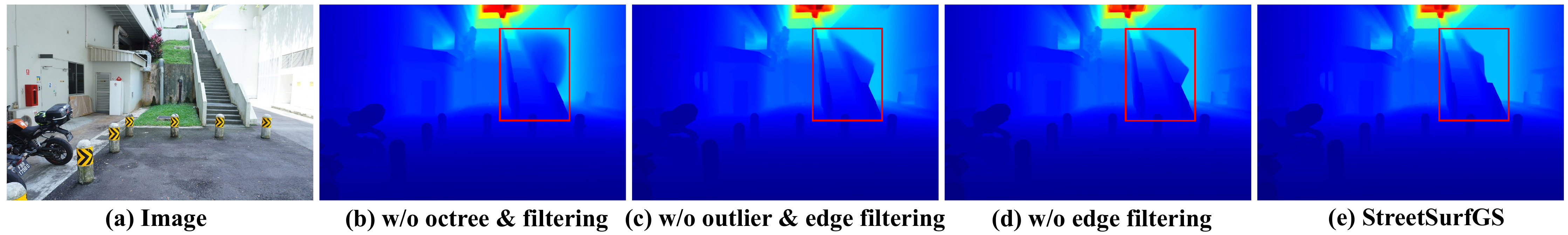}
\caption{Ablation study on depth map. The use of an octree structure and filtering strategies improves depth prediction accuracy and provides a more natural smoothness, leading to enhanced geometry information and, consequently, improved mesh quality.
}
\label{depth_stair}
\end{figure*}

\begin{figure*}[!ht]
\centering
\includegraphics[width=0.7\linewidth]{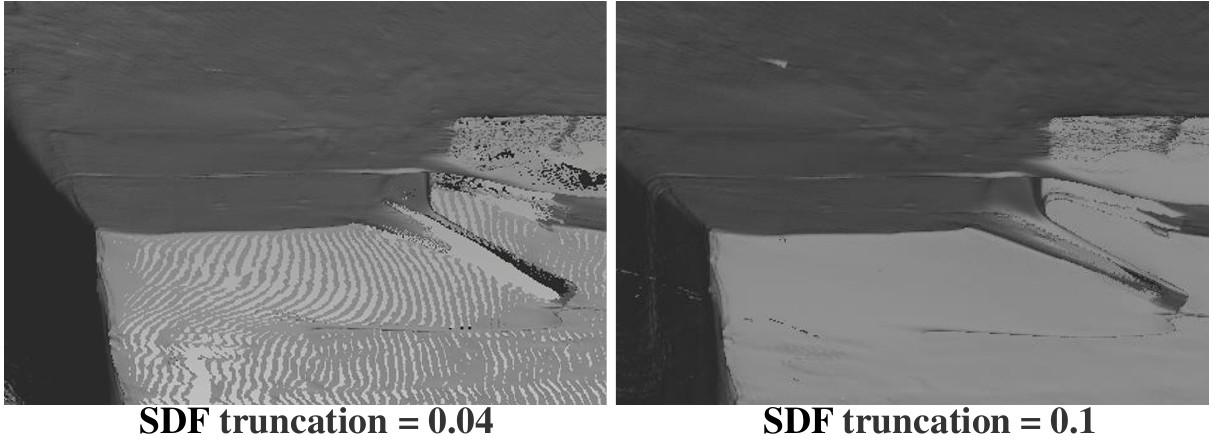}
\caption{Impact of SDF truncation in the TSDF fusion process. An appropriate SDF truncation value enhances the quality of surface reconstruction.
}
\label{sdftrunc}
\end{figure*}

%Given the inefficiency of 3DGS~\cite{kerbl20233d} and PGSR~\cite{chen2024pgsr} in handling large-scale scenes, making direct comparisons difficult, we provide results for applicable scenarios in the appendix.
% comparisons on all the scenes are difficult. We 

\subsection{Quantitative Comparisons}
% We conduct performance comparison of novel view synthesis using Free dataset~\cite{wang2023f2}, Waymo Open Dataset~\cite{sun2020scalability}, and KITTI-360~\cite{liao2022kitti}. As illustrated in Tables \ref{tab:kitti}, \ref{tab:waymo}, and \ref{tab:compare_free}, our approach demonstrates superior rendering performance, outperforming both surface reconstruction and purely rendering-focused methods. This improvement is primarily attributed to the octree structure, which enables fine-grained scene modeling, and the multi-view consistency guidance strategy, which effectively leverages information from both adjacent and distant frames, resulting in more robust modeling. 

% Given the inefficiency of 3DGS~\cite{kerbl20233d} and PGSR~\cite{chen2024pgsr} with large scenes, we cannot compare these methods across all scenes. Both methods are only capable of handling certain scenes from the KITTI-360 and Free datasets without exhausting memory resources. Therefore, we provide results on the KITTI-360 dataset in Table~\ref{tab:kitti} and a scene breakdown on the Free dataset in Table \ref{tab:compare_free_break_down}, reported in terms of PSNR. We also present memory usage in Table \ref{memory}. It is apparent that our method, compared to PGSR and 3DGS, not only provides higher rendering quality but also significantly reduces memory consumption. Our approach effectively considers the characteristics of street scenes and addresses associated challenges.

We conducted extensive performance comparisons for novel view synthesis across three datasets: the Free Dataset~\cite{wang2023f2}, the Waymo Open Dataset~\cite{sun2020scalability}, and KITTI-360~\cite{liao2022kitti}. Results summarized in Tables \ref{tab:kitti}, \ref{tab:waymo}, and \ref{tab:compare_free} exhibit our method's enhanced rendering capabilities, notably surpassing traditional surface reconstruction and existing rendering techniques. The substantial improvement can be largely attributed to our innovative use of an octree structure for fine-grained scene modeling, combined with a multi-view consistency guidance strategy. This strategy adeptly utilizes both adjacent and distant frame data, facilitating more accurate and robust scene renderings.

Due to the limitations of 3DGS~\cite{kerbl20233d} and PGSR~\cite{chen2024pgsr} in processing large scenes, direct comparisons across all scenarios were unfeasible, as both methods struggle with memory inefficiencies in expansive datasets. For feasible comparisons, we present specific results from the KITTI-360 dataset in Table~\ref{tab:kitti} and detailed scene-by-scene performance on the Free dataset in Table \ref{tab:compare_free_break_down}, with evaluations based on Peak Signal-to-Noise Ratio (PSNR). Additionally, memory consumption comparisons are detailed in Table \ref{memory}. Our method not only consistently outperforms PGSR and 3DGS in rendering quality but also demonstrates a significant reduction in memory usage. This efficiency is critical for handling the extensive and detailed street scenes typically encountered in these datasets.

The challenge of acquiring accurate ground-truth geometry in large-scale environments complicates the evaluation of geometry-focused metrics. To address this, we utilized the MatrixCity dataset~\cite{li2023matrixcity}, which provides reliable ground-truth geometrical data. We aligned the mesh generated by the baseline method with the points from COLMAP for better evaluation. Comparative results displayed in Table \ref{matri} confirm that our approach not only preserves geometric fidelity but also excels in geometry reconstruction, outperforming existing methods. This capability ensures that our surface reconstructions maintain integrity without sacrificing the quality of rendering, thus meeting both aesthetic and practical demands in visual applications.

\subsection{Qualitative Comparisons}

\textbf{Overview of Comparisons:}
We present a comprehensive qualitative analysis using the KITTI-360 dataset~\cite{liao2022kitti}, as depicted in Figure~\ref{mesh_kitti}. Detailed evaluations, including novel view synthesis, mesh reconstruction, and depth rendering, are conducted on both KITTI-360 and the Free Dataset, illustrated across Figures~\ref{v3}, \ref{render_free}, \ref{v1}, \ref{v2}, \ref{fig:mesh_free}, and \ref{v4}.

\textbf{Rendering Quality:}
Figures \ref{v3} and \ref{render_free} display the rendering results. Our method significantly enhances image clarity and captures intricate details. Previous approaches predominantly depend on immediate frame data, which often results in blurred object peripheries. In contrast, our method extends the data scope by leveraging adjacent frame information coupled with long-term spatial constraints, which sharply reduces such blurring effects.

\textbf{Surface Reconstruction:}
Figure \ref{v1} contrasts surface reconstruction outcomes on KITTI-360. Traditional techniques show reasonable performance in focused, object-centric scenes but underperform in dynamic, elongated environments like urban street views, producing only rudimentary outlines and noisy mesh textures. In stark contrast, our method not only delineates finer structural details but also ensures enhanced mesh fidelity. The qualitative superiority of our mesh reconstruction is further highlighted in Figure \ref{v2}, which features the textured meshes generated by our approach. Additionally, Figure \ref{fig:mesh_free} provides a comparative analysis of surface reconstructions using the Free Dataset, underscoring our technique's ability to create detailed and accurate meshes compared to existing methods like SuGaR~\cite{guedon2023sugar}.

\textbf{Depth Rendering Accuracy:}
As demonstrated in Figure \ref{v4}, previous methods exhibit insufficient depth details, particularly for distant objects, resulting in edge blurriness and poorly defined depth gradations. Our methodology addresses these deficiencies by incorporating long-term constraints that enhance far-field depth accuracy. This strategy allows for precise depth adjustments and adeptly models objects at varied distances, significantly improving the clarity and accuracy of depth maps.

\begin{figure}[!ht]
    \centering
        \includegraphics[width=0.6\linewidth]{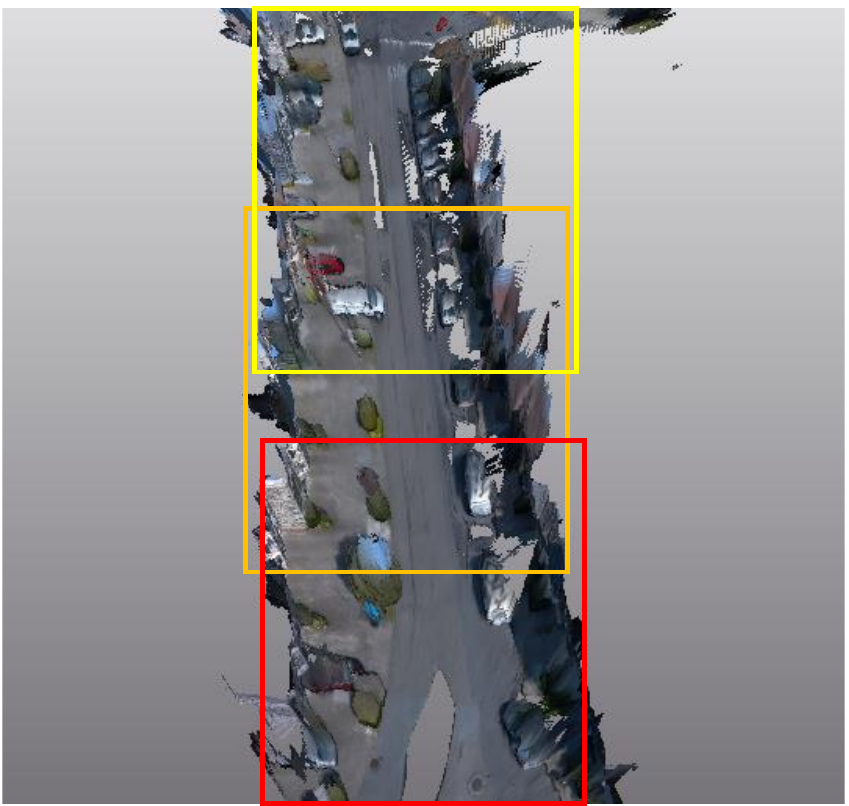}
        \caption{Illustration of our segmented training strategy. We divide the scene into a series of overlapping regions
that are reconstructed sequentially. The depth map from the
previously reconstructed segmented part is used to supervise
the subsequent part, providing a form of long-distance supervision that mitigates error accumulation. After reconstructing
all segments, we use all the images and depth maps to perform
mesh extraction.}
        \label{seg}
        \end{figure}
\begin{figure}
        \centering
        \includegraphics[width=0.8\linewidth]{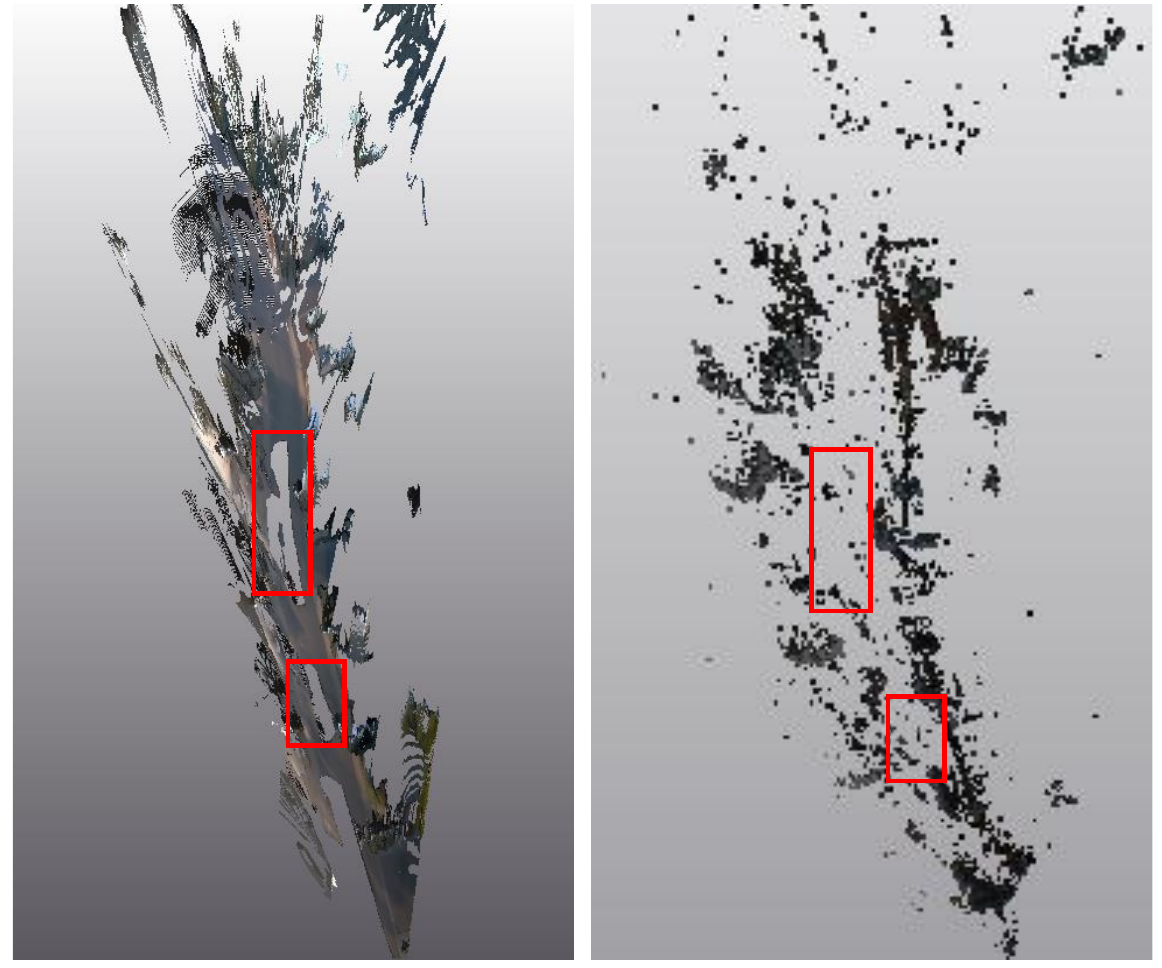}
        \caption{Failure case attributed to inaccurate initial point cloud. COLMAP structure-from-motion technique sometimes struggles to adequately model road surfaces, resulting in sparse point clouds, particularly in certain areas where there is almost no data available. Consequently, this leads to substantial gaps in the reconstructed road surface.}
        \label{failure}
\end{figure}

\subsection{Ablation Study}

We conduct ablation studies on the KITTI-360 and Free Dataset, the results of which are illustrated in Tables~\ref{ablation},~\ref{ablation2} and Figures~\ref{mesh_ablation},~\ref{depth_stair}. These studies evaluate the individual contributions of each component of our proposed system.

\textbf{Ablation of Octree Structure:}
The octree structure is designed to meticulously capture and preserve the structural fidelity of urban environments. Results presented in Tables~\ref{ablation} and~\ref{ablation2} highlight the octree’s capability for fine-grained scene modeling, significantly increasing rendering fidelity. As depicted in Figures~\ref{mesh_ablation} and~\ref{depth_stair}, the use of an octree enables more precise modeling and enhances the understanding of object positioning relationships, thus contributing to more accurate scene reconstructions.

\textbf{Ablation of Matching Strategies:}
Our matching strategies address challenges associated with sparse viewpoints typical of street view scenes and manage details over varying distances. The efficacy of our far-field matching strategy, as shown in Tables~\ref{ablation} and~\ref{ablation2}, is evident in its ability to significantly enhance rendering quality by introducing long-term constraints that improve both depth and positional accuracy across frames.

\textbf{Ablation of Filtering Strategies:}
The filtering strategies are designed to correct depth continuity errors and resolve complex spatial relationships between overlapping objects. The impact of these strategies on rendering quality is substantial, especially when the matching strategy provides sufficient contextual information. This is reflected in both tables and further illustrated in Figures~\ref{mesh_ablation} and~\ref{depth_stair}, where the integration of smoothness constraints, outlier filtering, and edge filtering markedly improves the accuracy of geometric reconstructions. These filters effectively manage object interactions and contribute to a more coherent and visually appealing output.

\subsection{Hyperparameter in TSDF Fusion}
\label{a4}
TSDF fusion is a commonly used technique in 3D reconstruction, which integrates depth information from multiple viewpoints to generate a continuous 3D surface representation. During this process, the SDF truncation size directly influences the accuracy, completeness, and memory consumption of surface reconstruction. A smaller truncation size allows for capturing more surface details but may result in discontinuities in the surface. Conversely, a larger truncation size can produce a more continuous surface but may sacrifice some detail, while also increasing the need to store additional voxels, thus increasing memory consumption. Therefore, selecting an appropriate truncation size is crucial. Our experiments suggest that setting the SDF truncation to 0.1 strikes a good balance between these factors. For a concrete example, please refer to Figure \ref{sdftrunc}, where an appropriate SDF truncation value significantly enhances the quality of surface reconstruction.

\subsection{Segmented Training}
% \label{a5}
Although we optimized memory usage with an octree structure, this does not completely solve the memory limitations for large-scale scenes. In long and narrow street scenes, information from frames captured far apart in time has minimal overlap, making it ineffective for supervision. To address this, we can partition the scene. Unlike previous NeRF-based methods that rely on marching cubes requiring the entire density field for reconstruction, TSDF fusion only needs input images, depth maps, and camera poses for reconstruction. This allows us to partition the scene during the modeling process. To ensure overall consistency in scene modeling and accurate alignment in TSDF fusion, as illustrated in Figure~\ref{seg}, we divide the scene into a series of overlapping regions that are reconstructed sequentially. The depth map from the previously reconstructed segmented part is used to supervise the subsequent part, providing a form of long-distance supervision that mitigates error accumulation. After reconstructing all segments, we use all the images and depth maps to perform mesh extraction through TSDF fusion. This segmented training strategy enables scalable large scene surface reconstruction.

\subsection{Failure Case}
\label{a6}
%我们使用colmap SfM产生初始稀疏点云。如图1所示，colmap有时候对路面无法形成很好的建模，在部分路面几乎没有点云。在这种情况下，我们有时候也无法在这里产生点云，造成重建的路面有大片空缺。
% Due to the lack of texture information on roads, some scenes produce sparse point clouds
% from COLMAP that are predominantly distributed at the sides rather than on the road surface itself,
% resulting in incomplete road surface reconstruction.
We use the COLMAP structure-from-motion technique to generate an initial sparse point cloud. As depicted in Figure~\ref{failure}, COLMAP sometimes struggles to adequately model road surfaces, resulting in sparse point clouds, particularly in certain areas where there is almost no data available. Consequently, this leads to substantial gaps in the reconstructed road surface. To address this limitation in future work, one could explore the use of enhanced algorithms within the structure-from-motion framework or consider alternative methods to generate more robust initial point cloud of road surfaces.

% We conduct comprehensive ablation studies to evaluate the impact of our proposed components on novel view synthesis and surface reconstruction tasks. As shown in Table~\ref{ablation}, our far-field matching strategy significantly enhances rendering performance. Furthermore, when combined with our filtering strategy, the modeling capabilities for novel view synthesis are further improved. Figure~\ref{mesh_ablation} and~\ref{depth_stair} illustrate the influence of different components on geometry reconstruction. The results demonstrate that our filtering strategy effectively addresses discontinuities between different objects, leading to superior performance in handling the sparse views and complex object arrangements characteristic of urban street scenes.
% \noindent\textbf{Octree-based architecture.}

% \noindent\textbf{Filtering strategies on smooth regularization.}

% \noindent\textbf{Multi-view consistency regularization.}

\section{Conclusion}
\label{conclusion}
We present the first exploration of employing Gaussian Splatting for scalable surface reconstruction in urban street scenes. Employing a planar-based octree Gaussian splatting representation, we implement a segmented training strategy to overcome memory constraints and ensure scalability. Within each segmented portion, we employ filtering strategies during smoothness constraints to mitigate depth continuity errors and employ matching strategies to integrate adjacent and long-term information. Extensive experiments confirm that our approach achieves superior rendering quality and delivers satisfactory surface reconstruction results.

% To address memory constraints, we adopt a planar-based octree Gaussian splatting representation and implement a segmented training strategy to ensure scalability. For each segmented portion, considering the elongated and narrow characteristics of street scenes, as well as challenges such as weak textures and sparse viewpoints common in such scenes, we employ filtering masks during smoothness constraints and leverage distant frames to establish long-term constraints. Extensive experiments demonstrate that our approach achieves superior rendering quality and produces satisfactory surface reconstruction results.

\noindent\textbf{Limitation.} Due to the lack of texture information on roads, some scenes produce sparse point clouds from COLMAP that are predominantly distributed at the sides rather than on the road surface itself, resulting in incomplete road surface reconstruction. 
%Specific examples are provided in the appendix.

\noindent\textbf{Broader Impact.} The meshes generated by our method can be used for object interaction in street-level scenes, offering convenience for autonomous driving and urban planning applications. Moreover, our approach can serve as a foundational model for large-scale reconstruction in various contexts. Additionally, our method is applicable to Virtual Reality (VR) and Augmented Reality (AR) scenarios.

% \section{Acknowledgements} 
% This work was partially supported by .

% \bibliographystyle{plain}
% \newpage
\bibliographystyle{IEEEtran}
\bibliography{reference}

\end{document}